\documentclass{article}

\usepackage[main, final]{neurips_2026}

\makeatletter

\renewenvironment{table}
  {\setlength{\abovecaptionskip}{\@neuripsbelowcaptionskip}%
   \setlength{\belowcaptionskip}{\@neuripsabovecaptionskip}%
   \@float{table}}
  {\end@float}

\renewenvironment{table*}
  {\setlength{\abovecaptionskip}{\@neuripsbelowcaptionskip}%
   \setlength{\belowcaptionskip}{\@neuripsabovecaptionskip}%
   \@dblfloat{table}}
  {\end@dblfloat}

\makeatother

\usepackage[utf8]{inputenc} 
\usepackage[T1]{fontenc}    
\usepackage{hyperref}       
\usepackage{url}            
\usepackage{booktabs}       
\usepackage{amsfonts}       
\usepackage{nicefrac}       
\usepackage{microtype}      

\usepackage{algorithm}
\usepackage{algpseudocode}
\usepackage{amssymb}
\usepackage{amsmath}
\usepackage{bm}

\usepackage{graphicx}
\usepackage{subcaption} 
\usepackage[table]{xcolor}

\usepackage{multirow}
\usepackage{tabularx}   
\usepackage{xcolor}     
\usepackage{amssymb}    

\usepackage{makecell}
\usepackage{enumitem}
\newcommand{\op}[1]{\textsc{#1}}
\definecolor{PosColor}{HTML}{1E9E5A}
\definecolor{NegColor}{HTML}{D1495B}
\definecolor{NeuColor}{HTML}{7A7A7A}

\newcommand{\poss}[1]{\textcolor{PosColor}{\ensuremath{+\mathbf{#1}}}}
\newcommand{\negg}[1]{\textcolor{NegColor}{\ensuremath{-\mathbf{#1}}}}
\newcommand{\neuu}{\textcolor{NeuColor}{\ensuremath{\mathbf{\pm0.0}}}}

\title{Dynamic Query as Budgeting for Tiny Object Detection}

\author{%
  \textbf{Yi Zhang\textsuperscript{1} \quad Di Xiong\textsuperscript{1} \quad Ge Gao\textsuperscript{1} \quad Xiangyue Zhang\textsuperscript{1}} \\
  \textbf{Yihang Qiu\textsuperscript{1} \quad Yuxuan Zhou\textsuperscript{1} \quad Shuo Chen\textsuperscript{1,2}\thanks{Corresponding author.}} \\
  \textsuperscript{1}School of Intelligence Science and Technology, Nanjing University, China \\
  \textsuperscript{2}Center for Advanced Intelligence Project, RIKEN, Japan \\
  \texttt{zhangyi\_magic@smail.nju.edu.cn}
  \quad \texttt{shuo.chen@nju.edu.cn}
}

\begin{document}

\maketitle

\begin{abstract}
Recent DETR-based tiny object detectors adopt dynamic query mechanisms to handle severe density imbalance. 
However, existing designs entangle two effects: how many and which queries are decoded. 
This entanglement obscures what drives the gains and complicates dynamic-query design; additional learned predictors and heuristic filtering also introduce overhead and potential selection errors.
We disentangle these effects and show that dynamic query primarily acts as \emph{budgeting}---allocating decoder query capacity according to object density---rather than intricate query selection.
Motivated by this view, we formulate a monotonic and conservative budgeting principle and propose a budgeting-based DETR (\textbf{BUTR}), a streamlined framework that uses a lightweight, non-learned budgeter to estimate an input-dependent query budget $K(x)$ from encoder confidences.
BUTR retains standard Top-$K(x)$ query initialization without additional learned modules or heuristic filtering in the budgeter.
At $800\times800$ resolution, BUTR-L achieves 35.7 AP at 47.8 FPS on AI-TOD-v2 and 29.8 AP at 42.1 FPS on VisDrone on an NVIDIA T4 GPU.
Inference-time integration on COCO further demonstrates its applicability to general scenario detection.
Code is available at \url{https://github.com/zhangyi-magic/BUTR}.
\end{abstract}

\section{Introduction}
\label{sec:intro}
DETR-based detectors frame object detection as set prediction, simplifying the pipeline by reducing hand-crafted heuristics~\cite{carion2020end, zhu2020deformable}.
Recent advances have improved convergence and efficiency, making this paradigm increasingly practical~\cite{zhao2024detrs, peng2024d, huang2025deim}.
Detecting tiny objects, however, remains challenging.
Tiny objects occupy only a few pixels, suffer from severe ambiguity, and are easily confused by background clutter.
In aerial imagery, this challenge is amplified by density imbalance: most images are sparsely populated, while a tiny fraction contains crowded scenes with hundreds (or more) instances~\cite{wang2021tiny, xu2022detecting, yang2022querydet, cheng2023towards}.
This creates a tension in DETR design: a fixed decoder query budget is efficient for the majority of images but may be insufficient for crowded cases, whereas uniformly increasing the budget imposes unnecessary overhead and can increase false positive detections in sparse scenes.

To address this, recent works introduce dynamic query mechanisms that adapt decoder capacity to each input.
DQ-DETR~\cite{huang2024dq} first predicts a density map and then uses a classifier to assign each image to a preset budget bin.
This two-stage discretized design is prone to error propagation, which can lead to unstable budget assignment across density scales.
In contrast, Dome-DETR~\cite{hu2025dome} adopts a selection-heavy procedure (Progressive Adaptive Query Initialization, PAQI) that combines a learned window-mask predictor with heuristic filtering, introducing sample-dependent control flow and multiple thresholds.
Although effective on challenging dense cases, these methods introduce extra predictors or heuristic operators, increasing complexity and reducing practicality without always improving accuracy.
More importantly, selection-heavy designs entangle query budgeting with query selection, obscuring the true source of improvement.

\begin{figure}[t]
\centering
\captionsetup{width=0.98\linewidth, font=small}
\begin{subfigure}[t]{0.32\linewidth}
    \centering
    \includegraphics[width=\linewidth]{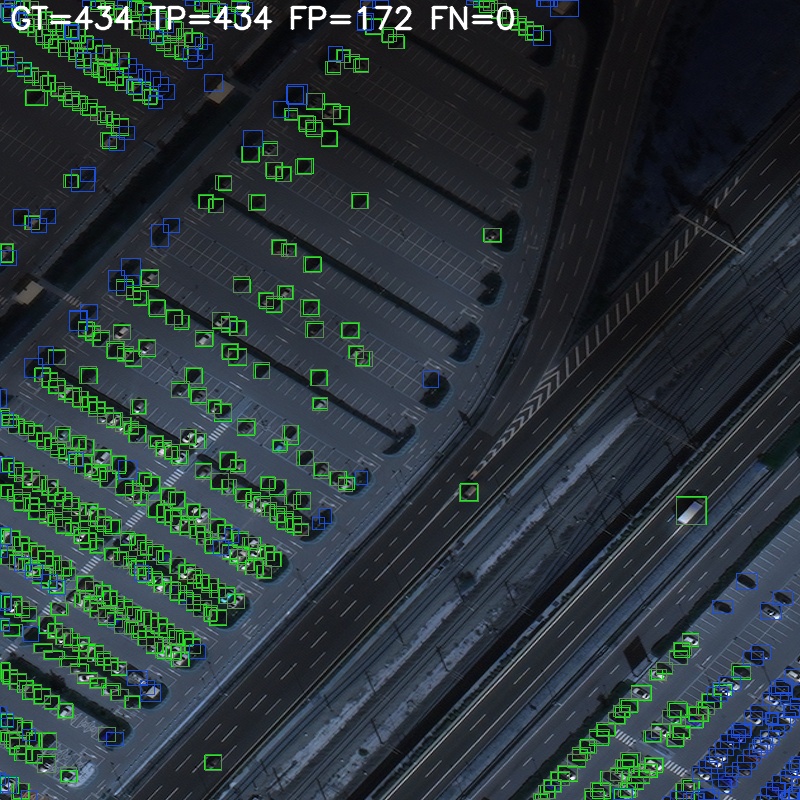}
    \vspace{-12pt}
    \caption{Dynamic}
    \label{fig:example2_dynamic}
\end{subfigure}
\begin{subfigure}[t]{0.32\linewidth}
    \centering
    \includegraphics[width=\linewidth]{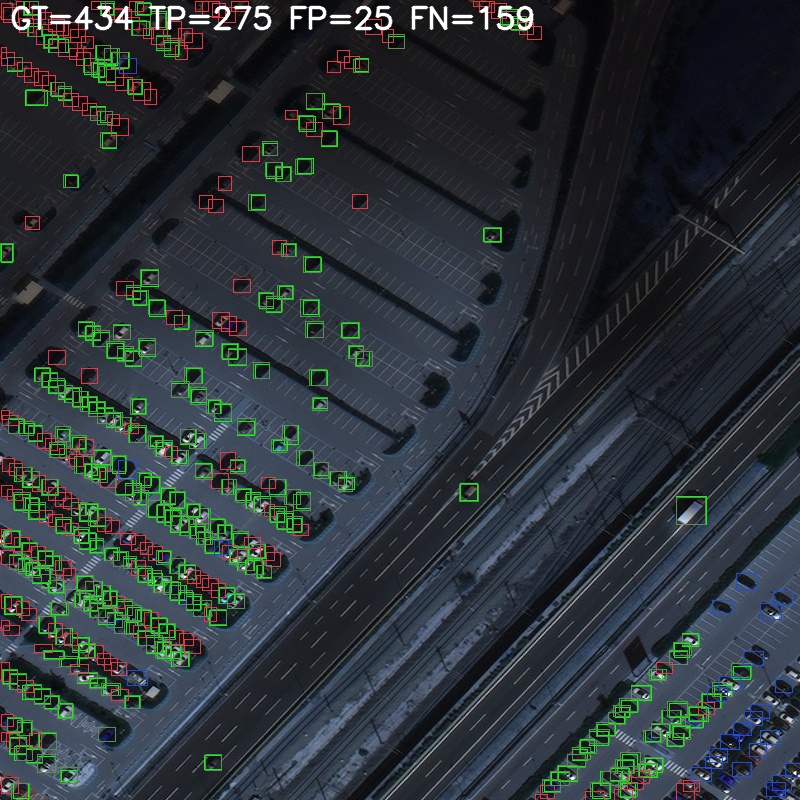}
    \vspace{-12pt}
    \caption{Fixed Top-300}
    \label{fig:example2_fixed600}
\end{subfigure}
\begin{subfigure}[t]{0.32\linewidth}
    \centering
    \includegraphics[width=\linewidth]{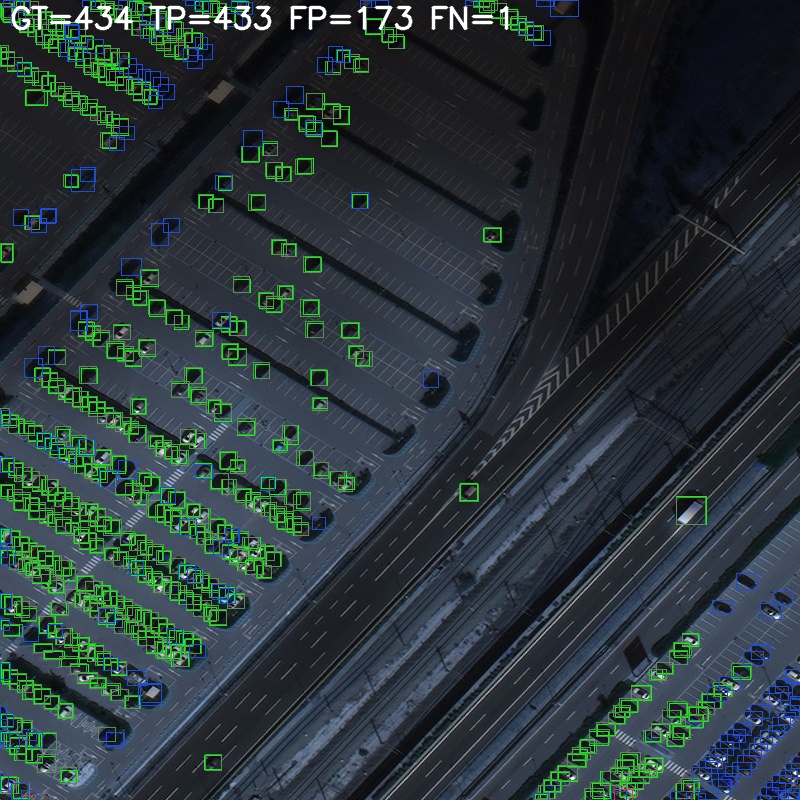}
    \vspace{-12pt}
    \caption{Fixed Top-600}
    \label{fig:example2_fixed1500}
\end{subfigure}
\caption{Qualitative comparison on AI-TOD-v2. Detection outcomes are color-coded as true positives (TP, green), false positives (FP, blue), and false negatives (FN, red). Replacing Dome-DETR's dynamic-query initialization in (a) with a fixed Top-$600$ initializer in (c) yields highly similar detection results, whereas the smaller fixed budget in (b) (Top-$300$) is clearly insufficient for this extreme-density scene.}
\label{fig:examples}
\end{figure}


In this paper, we revisit dynamic query from a mechanism-first perspective.
Existing designs conflate two effects: changing how many queries are decoded, and changing which queries survive.
To disentangle them, we analyze Dome-DETR~\cite{hu2025dome} through controlled inference-time analysis.
This leads to two key observations.
First, density-stratified evaluation reveals that, on AI-TOD-v2, PAQI's advantage over a fixed budget sufficient for the majority of images is concentrated on the rare high-density tail and largely obscured by overall AP.
Second, a simple confidence-ranked Top-$K$ initializer closely reproduces PAQI's accuracy not only in overall AP but also within density subsets when its fixed budget is adjusted to provide comparable query capacity; in particular, increasing $K$ also closes the gap on the high-density tail.
A qualitative example is shown in Fig.~\ref{fig:examples}.
Supplementary experiments with DDQ-style initialization~\cite{zhang2023dense} further support this interpretation: when the query count is held fixed, replacing confidence-ranked Top-$K$ with spatially distinct query selection yields the same overall AP on both AI-TOD-v2 and VisDrone (App.~\ref{app:ddq_style}).
Together, these findings suggest that dynamic query primarily acts as adaptive \emph{budgeting}---determining \emph{how many} queries to decode---rather than intricate selection of \emph{which} queries to keep.

Motivated by this budgeting perspective, we propose a budgeting-based DETR (BUTR), a streamlined framework that realizes dynamic query through a lightweight, non-learned budgeter that estimates a per-image query budget $K(x)$ from a simple density proxy over encoder proposals, while retaining standard Top-$K(x)$ initialization of two-stage DETRs.
Compared with prior dynamic-query detectors, BUTR delivers higher accuracy with lower overhead, faster training, and lower T4 deployment latency; as a drop-in modification for modern two-stage DETRs, it also makes dynamic query easier to implement, reproduce, and extend.
Our main contributions are summarized as follows:
\begin{itemize}[leftmargin=1.5em,itemsep=0.2em,topsep=0.2em]
\item \textbf{Perspective.} We disentangle how many and which queries are decoded, showing that dynamic query mainly performs adaptive decoder budgeting, with gains concentrated on the rare high density tail revealed by density-stratified evaluation.

\item \textbf{Method.} We formulate a budgeting principle for dynamic query and instantiate it as BUTR, which estimates a per-image decoder query budget $K(x)$ with a lightweight, non-learned budgeter to enable standard Top-$K(x)$ query initialization, avoiding extra selection machinery.

\item \textbf{Results.} On AI-TOD-v2 and VisDrone, BUTR achieves higher AP over prior dynamic-query detectors with lower overhead, and markedly faster training and lower T4 deployment latency; additional COCO results further indicate its applicability beyond tiny object detection.

\end{itemize}

\section{Related Work}
\label{sec:related_work}

\subsection{Fixed-Query DETRs}
DETR~\cite{carion2020end} formulates detection as end-to-end set prediction with a fixed set of object queries.
Subsequent variants have greatly improved convergence and efficiency, including deformable attention~\cite{zhu2020deformable}, query design and anchor-based query formulations~\cite{meng2021conditional,wang2022anchor,liu2022dab}, denoising or hybrid matching strategies~\cite{li2022dn,zhang2022dino,chen2023group,jia2023detrs}, and recent real-time architectures~\cite{zhao2024detrs,peng2024d,huang2025deim,zhang2023dense}, while retaining a fixed query budget (typically 300--900 queries).
On aerial tiny-object benchmarks such as VisDrone~\cite{zhu2021detection} and AI-TOD-v2~\cite{wang2021tiny, xu2022detecting}, fixed-query DETR-style detectors, together with related transformer-augmented two-stage baselines such as DNTR~\cite{liu2024denoising}, achieve strong performance.
However, under severe density imbalance, a fixed global query budget creates a natural trade-off: it is efficient for the majority of sparse images, but may become insufficient for rare crowded scenes.
This tension motivates recent efforts toward input-adaptive query allocation.

\subsection{Dynamic-Query DETRs}
To address this tension, recent works introduce dynamic queries that adapt decoder capacity to each input.
DQ-DETR~\cite{huang2024dq} adopts a learned strategy by predicting an image-level density signal and selecting among preset budget buckets.
Dome-DETR~\cite{hu2025dome}, by contrast, follows a selection-style strategy that filters a large candidate set through a multi-stage initialization pipeline (e.g., PAQI), yielding a variable-length decoder input.
Related adaptive-query ideas have also appeared in nearby remote-sensing settings, such as FDA-DETR~\cite{ju2025fda} for oriented small-object detection and CGAQ-DETR~\cite{zuo2025cgaq} for synthetic aperture radar object detection.
These methods highlight the promise of dynamic query, but also leave open a basic mechanism question: does it help mainly by changing how many queries are decoded, or by changing which queries are retained?

\section{Rethinking Dynamic Query: A Budgeting View}
\label{sec:diagnosis}
We analyze a strong recent dynamic-query method, Dome-DETR, as a diagnostic case, and decompose its query initialization behavior, since its explicit initialization pipeline, PAQI, makes the underlying mechanism directly testable.
PAQI starts from the Top-$K_{\max}$ encoder proposals and applies three operators:
(i) \textbf{\op{Split}}, which protects the Top-$K_{\min}$ proposals;
(ii) \textbf{\op{Mask}}, which filters the remaining proposals using a predicted window mask; and
(iii) \textbf{\op{NMS}}, which performs density-adaptive, class-aware suppression,
yielding variable-length decoder queries.
To isolate the role of each component and avoid retraining-induced confounders, we toggle these operators individually at inference-time.

\begin{table}[htbp]
\centering
\fontsize{8.5}{9.0}\selectfont
\setlength{\tabcolsep}{6pt}      
\captionsetup{width=0.98\linewidth, font=small} 
\caption{Inference-time ablation on AI-TOD-v2 and VisDrone. A Dome-DETR-L is trained on the respective \texttt{trainval} split and evaluated on the \texttt{test} set. For each configuration, we report query count $|Q|$ (mean $\mu$ $\pm$ std $\sigma$) and overall AP. $\Delta$ is relative to the full PAQI pipeline (Row~1).}
\vspace{2pt}
\label{tab:paqi_ablation}
\begin{tabular}{@{\hspace{2pt}}c@{\hspace{4pt}}|ccc@{\hspace{6pt}}cc@{\hspace{6pt}}cc}
\toprule
\multirow{2}{*}{Init.} &
\multicolumn{3}{c}{Configuration} &
\multicolumn{2}{c}{AI-TOD-v2} &
\multicolumn{2}{c}{VisDrone} \\
\cmidrule(lr){2-4} \cmidrule(lr){5-6} \cmidrule(lr){7-8}
 & Split & Mask & NMS & |Q| ($\mu\scriptstyle \, \pm \, \sigma$) & AP ($\Delta$) & |Q| ($\mu\scriptstyle \, \pm \, \sigma$) & AP ($\Delta$) \\
\midrule
Full PAQI & \checkmark & \checkmark & \checkmark & $630{\scriptstyle \, \pm \, 229}$ & 35.1 & $388{\scriptstyle \, \pm \, 21}$ & 29.3 \\
\cmidrule(lr){1-8}
\multirow{5}{*}{\makecell{Operator\\Variants}}
& \checkmark & \checkmark & $\times$   & $875{\scriptstyle \, \pm \, 395}$  & 35.1 (\neuu) & $494{\scriptstyle \, \pm \, 14}$ & 29.3 (\neuu) \\
& \checkmark & $\times$   & \checkmark & $1083{\scriptstyle \, \pm \, 116}$  & 35.1 (\neuu) & $449{\scriptstyle \, \pm \, 14}$ & 29.3 (\neuu) \\
& $\times$   & \checkmark & $\times$   & $759{\scriptstyle \, \pm \, 484}$  & 32.8 (\negg{2.3}) & $491{\scriptstyle \, \pm \, 23}$ & 29.2 (\negg{0.1}) \\
& $\times$   & \checkmark & \checkmark & $461{\scriptstyle \, \pm \, 299}$  & 32.7 (\negg{2.4}) & $299{\scriptstyle \, \pm \, 40}$ & 28.9 (\negg{0.4}) \\
& $\times$   & $\times$   & \checkmark & $895{\scriptstyle \, \pm \, 132}$  & 35.0 (\negg{0.1}) & $306{\scriptstyle \, \pm \, 36}$ & 29.0 (\negg{0.3}) \\
\cmidrule(lr){1-8}
\multirow{5}{*}{\makecell{Fixed \\ Top-$K$}}
& \multirow{5}{*}{$\times$} & \multirow{5}{*}{$\times$} & \multirow{5}{*}{$\times$}
& $100$  & 31.7 (\negg{3.4}) & $100$  & 27.0 (\negg{2.3}) \\
& & & & $300$  & 34.6 (\negg{0.5}) & $300$  & 29.1 (\negg{0.2}) \\
& & & & $600$  & 35.0 (\negg{0.1}) & $400$  & 29.3 (\neuu) \\
& & & & $900$  & 35.1 (\neuu) & $600$  & 29.4 (\poss{0.1}) \\
& & & & $1500$  & 35.1 (\neuu) & $1200$  & 29.5 (\poss{0.2}) \\
\bottomrule
\end{tabular}
\end{table}


\textbf{From selection to budgeting.}
Tab.~\ref{tab:paqi_ablation} reveals a clear contrast: removing \op{Split} causes substantial AP drops (Rows~4--6), whereas removing either \op{Mask} or \op{NMS} while retaining \op{Split} leaves overall AP unchanged despite varying query counts (Rows~1--3).
This contrast highlights the key role of \op{Split}: preserving the Top-$K_{\min}$ proposals, including high-confidence foreground proposals that the filtering operators may otherwise discard (see App.~\ref{app:mask_filtering} for examples).
PAQI can therefore be simplified to confidence-ranked proposal retention, naturally degenerating into a plain Top-$K$ initializer.

We then sweep the fixed budget $K$.
The Fixed Top-$K$ rows (Rows~7--11) show that a fixed budget close to PAQI's average query count closely reproduces overall accuracy on both datasets (AI-TOD-v2: AP $35.1$ with mean $|Q|{=}630$ vs. AP $35.0$ with fixed $K{=}600$; VisDrone: AP $29.3$ with mean $|Q|{=}388$ vs. AP $29.3$ with fixed $K{=}400$). Here, budget matching refers to approximate matching of the average query count, not an exact per-image match.
This indicates that once the decoder receives a sufficient number of high-confidence proposals, PAQI contributes little to overall AP.
App.~\ref{app:topk_overlap} provides two complementary explanations: PAQI already has high overlap with the confidence-ranked Top-$|Q|$ proposals, with a mean overlap of $71.5\%\pm 8.1\%$ on AI-TOD-v2 and $84.6\%\pm 1.9\%$ on VisDrone; and encoder proposals exhibit substantial redundancy around foreground objects.

\begin{table}[htbp]
\centering
\captionsetup{width=0.98\linewidth, font=small}
\fontsize{8.5}{9.0}\selectfont
\setlength{\abovecaptionskip}{0pt}
\setlength{\belowcaptionskip}{2pt}
\caption{Coarse density-stratified evaluation on AI-TOD-v2 and VisDrone.}
\vspace{2pt}
\label{tab:dynamic_query_degen}
\setlength{\tabcolsep}{1.5pt}
\renewcommand{\arraystretch}{0.9}
\begin{tabular}{c cc cc cc cc}
\toprule
\multirow{4}{*}{Init.} &
\multicolumn{4}{c}{AI-TOD-v2} &
\multicolumn{4}{c}{VisDrone} \\
\cmidrule(lr){2-5} \cmidrule(lr){6-9}
& \multicolumn{2}{c}{\texttt{Non-extreme}} & \multicolumn{2}{c}{\texttt{Extreme}} &
\multicolumn{2}{c}{\texttt{Non-dense}} & \multicolumn{2}{c}{\texttt{Dense}} \\
\cmidrule(lr){2-3} \cmidrule(lr){4-5} \cmidrule(lr){6-7} \cmidrule(lr){8-9}
& $|Q|$ ($\mu\scriptstyle\,\pm\,\sigma$) & AP ($\Delta$) & $|Q|$ ($\mu\scriptstyle\,\pm\,\sigma$) & AP ($\Delta$) &
$|Q|$ ($\mu\scriptstyle\,\pm\,\sigma$) & AP ($\Delta$) & $|Q|$ ($\mu\scriptstyle\,\pm\,\sigma$) & AP ($\Delta$) \\
\midrule
PAQI & $629{\scriptstyle\,\pm\,227}$ & 35.1 & $1326{\scriptstyle\,\pm\,100}$ & 29.9 &
$388{\scriptstyle\,\pm\,22}$ & 29.1 & $386{\scriptstyle\,\pm\,14}$ & 34.6 \\
\cmidrule(lr){1-9}
\multirow{2}{*}{Fixed Top-$K$}
& 600 & 35.1 (\neuu) & 600 & 23.4 (\negg{6.5}) &
\multirow{2}{*}{400} & \multirow{2}{*}{29.1 (\neuu)} & \multirow{2}{*}{400} & \multirow{2}{*}{34.5 (\negg{0.1})} \\
& 1500 & 35.1 (\neuu) & 1500 & 29.9 (\neuu) & & & & \\
\bottomrule
\end{tabular}
\end{table}

\textbf{Where does dynamic query matter?}
Tab.~\ref{tab:paqi_ablation} shows a clear saturation effect: once a fixed $K$ budget is sufficiently large, further increasing $K$ barely changes the overall AP.
This can be misleading under heavy density imbalance: a larger query count is expected to improve recall on dense images, yet this benefit may be diluted in the aggregate metric.
We therefore perform a coarse density-stratified analysis using thresholds of $\#\mathrm{GT}{=}600$ on AI-TOD-v2 and $\#\mathrm{GT}{=}300$ on VisDrone, where GT denotes ground-truth objects.
These thresholds are chosen near the observed saturation points with acceptable annotation counts, yielding \texttt{non-extreme}/\texttt{extreme} and \texttt{non-dense}/\texttt{dense} splits, respectively.
On AI-TOD-v2, this gives 13{,}992 \texttt{non-extreme} images (99.81\%) with 351{,}429 annotations (93.44\%) versus 26 \texttt{extreme} images (0.19\%) with 24{,}692 annotations (6.56\%); on VisDrone, it gives 1{,}603 \texttt{non-dense} images (99.57\%) with 72{,}588 annotations (96.66\%) versus 7 \texttt{dense} images (0.43\%) with 2{,}514 annotations (3.34\%).

As shown in Tab.~\ref{tab:dynamic_query_degen}, on AI-TOD-v2 PAQI behaves like Top-$600$ on the \texttt{non-extreme} majority, while its gains are confined to the rare \texttt{extreme} tail, where it increases the budget to $1326\pm100$; once the fixed budget is raised accordingly, Top-$1500$ closes the gap.
In contrast, VisDrone has a much weaker tail (max $\#\mathrm{GT}{=}461$): PAQI's query count barely changes across the split ($388\pm22$ vs.\ $386\pm14$), and a budget-matched fixed baseline (Top-$400$) reproduces both subsets.
Thus, the benefit of PAQI is confined to the high-density tail and is fully reproducible with a matched fixed budget.

\begin{figure}[htbp]
\centering
\captionsetup{width=0.98\linewidth, font=small}
\begin{subfigure}[t]{0.48\linewidth}
    \centering
    \includegraphics[width=\linewidth]{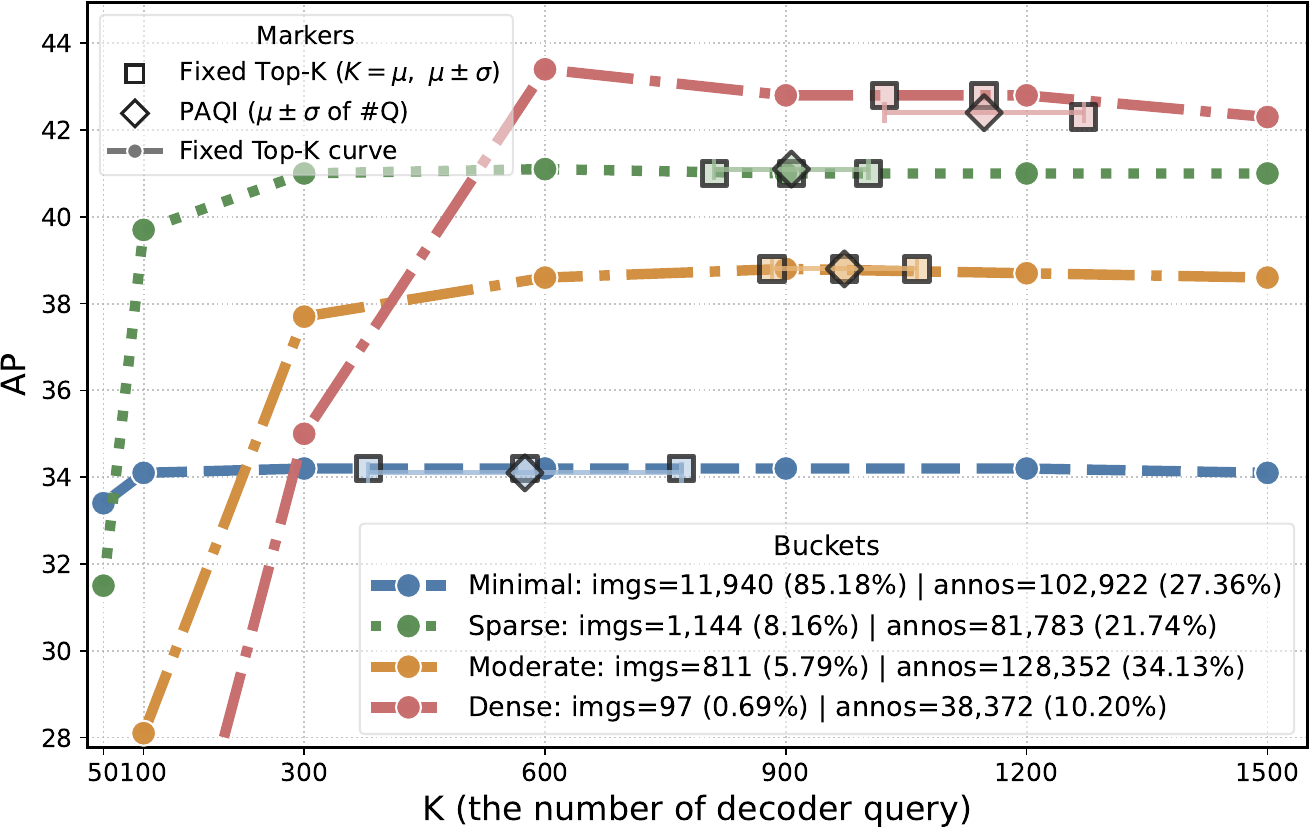}
    \caption{AI-TOD-v2}
    \label{fig:aitodv2_map_vs_k_stratified}
\end{subfigure}
\hspace{2pt}
\begin{subfigure}[t]{0.48\linewidth}
    \centering
    \includegraphics[width=\linewidth]{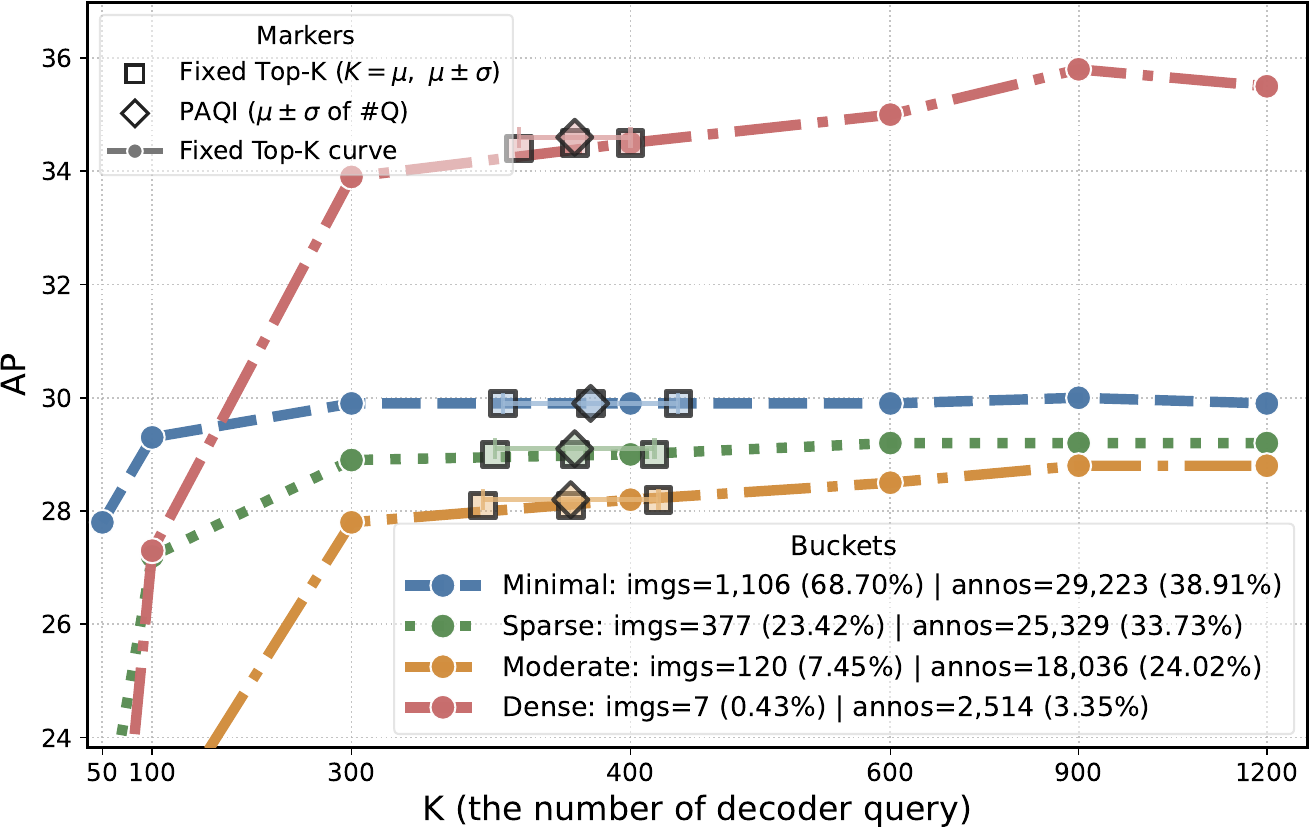}
    \caption{VisDrone}
    \label{fig:visdrone_map_vs_k_stratified}
\end{subfigure}
\caption{Fine density-stratified AP--budget response. Curves show AP versus fixed decoder budget $K$ within each density bucket: \textit{Minimal} ($\#\mathrm{GT}\le50$), \textit{Sparse} ($50<\#\mathrm{GT}\le100$), \textit{Moderate} ($100<\#\mathrm{GT}\le300$), and \textit{Dense} ($300<\#\mathrm{GT}\le600$). Diamonds mark PAQI's bucket-wise mean query count $\mu$ with horizontal bars for $\pm\sigma$; squares denote budget-matched fixed Top-$K$ baselines at $K\in\{\mu-\sigma,\mu,\mu+\sigma\}$.}
\label{fig:map_vs_k_stratified}
\end{figure}

\textbf{Budget matching reproduces density-wise accuracy.}
The finer bucket-wise AP--$K$ curves in Fig.~\ref{fig:map_vs_k_stratified} further support this interpretation.
Across density buckets, PAQI consistently lies near the response curve of a budget-matched Top-$K$ baseline, indicating that its behavior is largely explained by adaptive budgeting.
Moreover, AP quickly enters a saturation regime as $K$ increases, while PAQI often allocates more queries than needed, incurring extra compute with little or no accuracy gain.
This broad saturation regime explains why a simple density proxy suffices for dynamic budgeting: moderate estimation error rarely moves the operating point out of the flat region.

\textbf{Further evidence for the budgeting interpretation.}
The preceding analyses suggest that the primary role of dynamic query is to adapt decoder capacity to the object density of each image rather than to perform intricate query selection.
If this interpretation holds, changing \emph{which} queries are retained while keeping \emph{how many} fixed should yield nearly unchanged performance.
We test this prediction using DDQ-style initialization~\cite{zhang2023dense}, which retains a fixed-size, spatially distinct subset of candidates.
We separately train D-FINE-L with DDQ-style and original Top-$K$ initialization, using each initializer during both training and inference at the same query count; the two variants achieve the same overall AP on both AI-TOD-v2 and VisDrone.
App.~\ref{app:ddq_style} provides the motivation, rationale for choosing D-FINE, and detailed results.
Together with the PAQI interventions, these results support a mechanism-first interpretation: adaptive decoder budgeting accounts for much of the observed benefit of dynamic query in the evaluated settings, motivating BUTR's minimal design.

\section{BUTR}
\label{sec:BUTR}

\subsection{Preliminaries}

\begin{figure}[t]
\centering
\captionsetup{width=0.98\linewidth, font=small}
\includegraphics[width=1.0\linewidth]{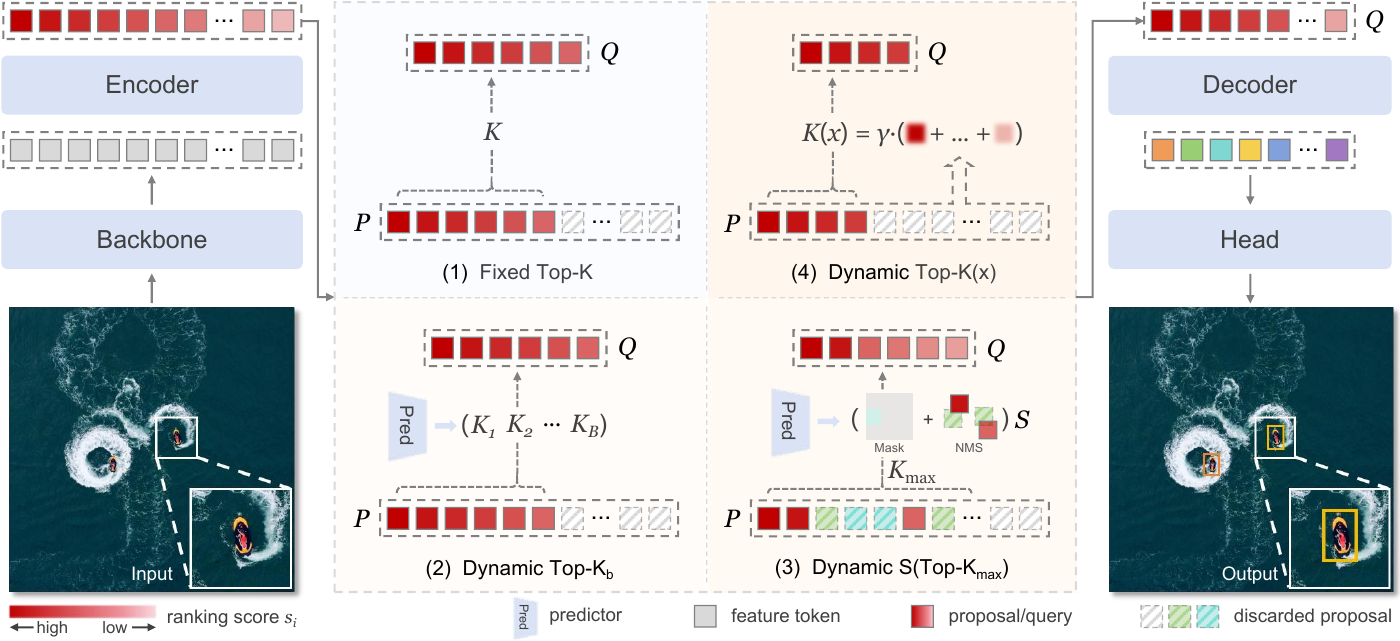}
\caption{Overview of query initialization: (1)~fixed-query DETRs (e.g., Deformable DETR~\cite{zhu2020deformable}, DINO~\cite{zhang2022dino}, and RT-DETR~\cite{zhao2024detrs}); dynamic-query DETRs include (2)~DQ-DETR~\cite{huang2024dq}, (3)~Dome-DETR~\cite{hu2025dome}, and (4)~ BUTR.}
\label{fig:models}
\end{figure}

\textbf{A unified view of query initialization.}
For clarity, we summarize query initialization in two-stage DETRs under a unified notation, as illustrated in Fig.~\ref{fig:models}.
Given an input image $x$, let $P=\{(\bm{z}_i,\bm{r}_i,\bm{\ell}_i)\}_{i=1}^{L}$
denote the set of encoder proposals, where $\bm{z}_i\in\mathbb{R}^{D}$ is the $i$-th proposal feature (or learned embedding), $\bm{r}_i$ is its reference box, and $\bm{\ell}_i=(\ell_{i1},\dots,\ell_{iC})\in\mathbb{R}^{C}$ is its classification logit vector, where $\ell_{ic}$ is the logit for class $c$.
Let $\mathcal{T}_{K}(\cdot)$ denote the Top-$K$ truncation operator based on a proposal ranking score $s_i=\phi(\bm{\ell}_i)$.
Standard two-stage DETRs (e.g., Deformable DETR~\cite{zhu2020deformable}, DINO~\cite{zhang2022dino}, and RT-DETR~\cite{zhao2024detrs}) use a fixed budget: $Q=\mathcal{T}_{K}(P)$ with constant $K$.
By contrast, dynamic-query variants adopt different strategies:
DQ-DETR~\cite{huang2024dq} performs discrete learned binning, $Q=\mathcal{T}_{K_b}(P)$ where $K_b\in\{K_1,\dots,K_B\}$;
selection-style methods such as Dome-DETR~\cite{hu2025dome} apply an additional selection pipeline to an enlarged candidate pool, $Q=\mathcal{S}\bigl(\mathcal{T}_{K_{\max}}(P)\bigr)$, with $\mathcal{S}(\cdot)$ denoting the selection pipeline.
Our BUTR instead keeps the standard Top-$K$ initialization form and only makes the truncation point adaptive based on proposal confidences: $Q=\mathcal{T}_{K(x)}(P)$, where $K(x)$ is a per-image budget.
This design introduces neither learned discrete predictors for $K_b$ nor a post-Top-$K$ selection pipeline $\mathcal{S}$, making BUTR a drop-in modification for two-stage DETRs.

\subsection{Budgeter: Estimating an Adaptive Query Budget}

\textbf{Budgeting principle.}
Following the diagnosis in Sec.~\ref{sec:diagnosis}, we identify that the essential role of dynamic query is to determine \emph{how many} decoder queries are needed for a given input.
Formally, we view a budgeter as a policy $\mathcal{B}:\mathcal{X}\rightarrow\mathcal{K}$ that assigns each input image $\bm{x}\in\mathcal{X}$ a feasible decoder budget.
A reasonable budgeting policy should satisfy two properties:
(i) \emph{monotonicity}---denser images should not receive fewer queries; and
(ii) \emph{conservativeness}---the policy should be mildly biased toward sufficient capacity, as under-allocation is more harmful than slight over-allocation in the observed AP--budget saturation regime (Fig.~\ref{fig:map_vs_k_stratified}).
Under this view, budgeting does not require recovering the exact object count, but only preserving a reliable ordering of image density and translating this ordering into an appropriately conservative query budget.

In BUTR, we provide a minimal implementation through a factorization $\mathcal{B}=g\circ e$, where
$e:\mathcal{X}\to\mathbb{R}_{+}, g:\mathbb{R}_{+}\to\mathcal{K}.$
Here, the estimator $e$ maps an input image to a nonnegative density-related signal, while the mapping $g$ converts this signal into a feasible decoder budget.
Under this decomposition, the estimator is responsible for preserving density ordering, while the mapping determines how conservatively this ordering is translated into decoder capacity.

\textbf{Estimator: a density proxy from encoder confidences.}
While the true object count is unavailable in practice, the estimator $e$ should preserve density ordering of inputs.
Fortunately, the ranking score $s_i=\phi(\boldsymbol{\ell}_i)$ already provides a natural and cost-free signal: it quantifies how likely the $i$-th encoder proposal carries foreground evidence.
Intuitively, if more proposals exhibit high scores, the image is likely to be denser and thus require more decoder queries.
We therefore define the density proxy as the sum of all ranking scores:
\begin{equation}
e(x)=\sum_{i=1}^{L}s_i=\sum_{i=1}^{L}\phi(\boldsymbol{\ell}_i).
\label{eq:density_proxy}
\end{equation}
This aggregation naturally respects the desired monotonicity.
In this work, we instantiate $\phi$ with a simple yet effective choice:
\begin{equation}
\phi(\boldsymbol{\ell}_i)
=
\max_{c\in\{1,\dots,C\}} \sigma(\ell_{ic})
\in [0,1],
\label{eq:si_def}
\end{equation}
where $\sigma(u)=\frac{1}{1+e^{-u}}$ is the sigmoid function.
That is, $\phi(\boldsymbol{\ell}_i)$ is the maximum sigmoid-activated class confidence across all classes, which we use as the proposal-level foreground confidence.
This choice is compatible with established detection practice, where confidence or quality-aware scores serve as generic indicators of foreground evidence in proposal generation and dense detection~\cite{ren2016faster,lin2017focal,tian2020fcos,zhang2020bridging,li2020generalized,zhang2021varifocalnet}.
Other instantiations of $\phi$ are possible and are examined in the ablation study (Sec.~\ref{sec:exp_ablation}).

\textbf{Mapping: from density proxy to decoder budget.}
Given the estimator output $e(x)$, budgeting requires a mapping $g$ that converts this density proxy into a decoder budget.
This mapping should naturally be monotone and also encode conservativeness.
Other monotone choices are also possible, such as discrete binning functions; in this work, we adopt a simple linear scaling.
Specifically, we instantiate the mapping as $g(t)=\gamma t, \gamma\ge 1$, yielding the budget:
\begin{equation}
K(x)=\gamma\cdot e(x).
\label{eq:budget_rule}
\end{equation}
Here, $\gamma$ is a safety factor controlling how conservatively the density signal is translated into budget. Alg.~\ref{alg:butr} in App.~\ref{app:butr_algorithm} summarizes the complete dynamic-query initialization procedure of BUTR.

\textbf{Empirical properties of the estimator.}
Empirically, $e(x)$ correlates strongly with the true object count (AI-TOD-v2: Pearson $r=0.94$, Spearman $\rho=0.82$; VisDrone: $r=0.82$, $\rho=0.79$; all $p$-values $<10^{-5}$), supporting its use as a density-related budgeting signal on these datasets.

To summarize, BUTR provides a simple, non-learned budgeter with negligible overhead: it introduces no learnable parameters, auxiliary supervision, additional loss, or budgeter-specific optimization. The host detector is still trained using its detection objectives.
By preserving the standard two-stage initialization form, it remains fully compatible with existing two-stage DETR pipelines and can even serve as a plug-in modification without retraining.

\section{Experiments}
\label{sec:experiments}

\subsection{Setup}
\label{sec:exp_data}
\textbf{Datasets and metrics.}
Our main comparisons use AI-TOD-v2~\cite{wang2021tiny, xu2022detecting} and VisDrone~\cite{zhu2021detection}, two widely used tiny-object benchmarks with stronger density variation than COCO~\cite{lin2014microsoft}. Sec.~\ref{sec:generality} additionally evaluates COCO to assess applicability beyond tiny-object detection.
We report AP, cost, training time, deployment latency, and the average number of decoder queries ($|Q|$), and additionally adopt density-stratified evaluation to expose high-density-tail behavior.

\textbf{Fair comparison protocol.}
For the main comparisons, we set the optional feasible range $[K_{\min}, K_{\max}]$ in Alg.~\ref{alg:butr} to match Dome-DETR's~\cite{hu2025dome} decoder query budget ranges: $[300,1500]$ on AI-TOD-v2 and $[250,500]$ on VisDrone. The common upper bound rules out a larger permitted maximum query budget as the source of accuracy gains, while the common lower bound rules out a smaller permitted minimum query budget as the source of efficiency gains. These ranges serve the comparison protocol rather than define the budget estimator; Fig.~\ref{fig:budget_vs_gt} separately examines its behavior without dataset-specific clipping.
DQ-DETR~\cite{huang2024dq} is reported under a 36-epoch schedule on both datasets.
For fixed-query baselines, we adapt them to the tiny and dense regime by using 600/400 queries on AI-TOD-v2/VisDrone, following the saturation points diagnosed in Sec.~\ref{sec:diagnosis}, and by extending training to 160/120 epochs to match Dome-DETR and BUTR, with epoch-based schedules scaled proportionally (e.g., data-augmentation milestones).
Optimization and objective settings (e.g., learning rate and loss-related hyperparameters) are aligned with each dataset; for DEIM, we replace its Matchability-Aware Loss (MAL)~\cite{huang2025deim} with Varifocal Loss~\cite{zhang2021varifocalnet}, since MAL is unstable in our setting and leads to collapsed AP.
For these tiny-object comparisons, BUTR, D-FINE, DEIM, and Dome-DETR use four feature levels with an anchor size of 0.0125; DQ-DETR retains its five-level architecture. Models are trained on the \texttt{trainval} split and evaluated on the \texttt{test} split.
Following Dome-DETR, the L-scale configurations use four decoder layers on AI-TOD-v2 and six on VisDrone, accounting for their different parameter counts.

\textbf{Implementation details.}
\label{sec:exp_impl}
Training uses a server with two 7R32 CPUs and eight RTX 4090 GPUs, with $800\times800$ inputs. Training time is measured with total batch size 16 and per-epoch validation.
Deployment latency is measured on a single NVIDIA Tesla T4 (16\,GB) with TensorRT 10.4.0, FP16 enabled, and batch size 1; precision constraints and the measurement protocol are detailed in App.~\ref{app:t4_deployment}.

\subsection{Comparisons}
\label{sec:exp_main}

\begin{table}[htbp]
\centering
\fontsize{8.2}{8.8}\selectfont
\captionsetup{width=0.98\linewidth, font=small}
\setlength{\tabcolsep}{1.2pt}
\renewcommand{\arraystretch}{1.12}
\caption{Comparison on AI-TOD-v2. DQ-DETR results use the official checkpoint; others are our re-implementations. Dynamic-query GFLOPs and $|Q|$ are averages. Tr. (h): training time (hours); Lat. (ms): T4 TensorRT latency (milliseconds). Ratios are computed relative to Dome-DETR at the same model scale.}
\vspace{2pt}
\label{tab:aitodv2_results}
\begin{tabular}{l|ccccc|ccccccc}
\toprule
\textbf{Method} & \textbf{\#Params} & \textbf{GFLOPs} & $\boldsymbol{|Q|}$ & \textbf{Tr. (h)} & \textbf{Lat. (ms)} & \textbf{AP} & \textbf{AP}$_{50}$ & \textbf{AP}$_{75}$ & \textbf{AP}$_{vt}$ & \textbf{AP}$_t$ & \hspace{2pt}\textbf{AP}$_s$ & \textbf{AP}$_m$ \\
\midrule

\rowcolor{orange!8}
\multicolumn{13}{c}{\textbf{Fixed-query detectors}} \\
D-FINE-S~\cite{peng2024d} & 11.6M & 115.1 & 600 & 33.7 & 11.27 & 32.1 & 65.5 & 27.4 & 17.5 & 32.3 & 38.1 & 45.7 \\
D-FINE-M~\cite{peng2024d} & 22.3M & 248.2 & 600 & 40.4 & 16.86 & 34.3 & 68.0 & 30.3 & 19.7 & 34.8 & 39.1 & 47.7 \\
D-FINE-L~\cite{peng2024d} & 31.9M & 329.6 & 600 & 43.0 & 20.34 & 35.2 & 69.3 & 31.1 & 20.4 & 36.0 & 40.0 & 47.9 \\
DEIM-S~\cite{huang2025deim} & 11.6M & 115.1 & 600 & 36.6 & 11.26 & 29.0 & 61.0 & 23.9 & 16.8 & 28.6 & 35.0 & 41.7 \\
DEIM-M~\cite{huang2025deim} & 22.3M & 248.2 & 600 & 43.4 & 16.92 & 33.7 & 67.3 & 30.1 & 20.3 & 34.3 & 38.4 & 46.1 \\
DEIM-L~\cite{huang2025deim} & 31.9M & 329.6 & 600 & 46.6 & 20.34 & 34.5 & 68.5 & 30.7 & 20.4 & 35.0 & 38.9 & 46.5 \\
\midrule

\rowcolor{orange!8}
\multicolumn{13}{c}{\textbf{Dynamic-query detectors}} \\
DQ-DETR~\cite{huang2024dq} & 58.7M & 1805.4 & 410 & 14.8 & 205.58 & 30.5 & 69.2 & 22.7 & 15.2 & 30.9 & 36.8 & 45.5 \\
Dome-DETR-S~\cite{hu2025dome} & 13.2M & 165.3 & 639 & 65.0 & 23.77 & 32.3 & 67.0 & 27.5 & 18.2 & 32.4 & 38.7 & 45.6 \\
Dome-DETR-M~\cite{hu2025dome} & 23.9M & 298.5 & 632 & 78.9 & 30.01 & 34.0 & 68.4 & 29.8 & 19.6 & 34.3 & 39.5 & 48.3 \\
Dome-DETR-L~\cite{hu2025dome} & 33.4M & 379.8 & 630 & 83.6 & 33.30 & 35.1 & 69.6 & 30.8 & 19.2 & 35.6 & 39.9 & 47.7 \\
\cmidrule(lr){1-13}
\textbf{BUTR-S} (Ours) & 11.6M & 112.5 & 319 & 30.8\hspace{1pt}\fontsize{7.8}{9}\selectfont(\textcolor{PosColor}{2.1$\times$}) & 12.23\hspace{1pt}\fontsize{7.8}{9}\selectfont(\textcolor{PosColor}{1.9$\times$}) & \textbf{32.6}\hspace{1pt}\textbf{\fontsize{7.8}{9}\selectfont(\textcolor{PosColor}{$+$0.3})} & 67.2 & 27.8 & 19.0 & 32.2 & 38.6 & 45.4 \\
\textbf{BUTR-M} (Ours) & 22.3M & 244.8 & 317 & 37.2\hspace{1pt}\fontsize{7.8}{9}\selectfont(\textcolor{PosColor}{2.1$\times$}) & 17.60\hspace{1pt}\fontsize{7.8}{9}\selectfont(\textcolor{PosColor}{1.7$\times$}) & \textbf{34.6}\hspace{1pt}\textbf{\fontsize{7.8}{9}\selectfont(\textcolor{PosColor}{$+$0.6})} & 69.2 & 30.5 & 20.3 & 34.7 & 38.9 & 47.2 \\
\textbf{BUTR-L} (Ours) & 31.9M & 326.1 & 316 & 39.8\hspace{1pt}\fontsize{7.8}{9}\selectfont(\textcolor{PosColor}{2.1$\times$}) & 20.94\hspace{1pt}\fontsize{7.8}{9}\selectfont(\textcolor{PosColor}{1.6$\times$}) & \textbf{35.7}\hspace{1pt}\textbf{\fontsize{7.8}{9}\selectfont(\textcolor{PosColor}{$+$0.6})} & 69.7 & 32.2 & 21.2 & 36.3 & 40.6 & 49.3 \\
\bottomrule
\end{tabular}
\end{table}

\textbf{AI-TOD-v2.}
Tab.~\ref{tab:aitodv2_results} shows that a strong fixed-query baseline, D-FINE, already surpasses Dome-DETR in overall AP. One plausible reason is that Dome-DETR mistakenly discards high-confidence proposals aligned with foreground objects; App.~\ref{app:mask_filtering} provides qualitative examples. This result further supports our budgeting view.
Compared with fixed-query baselines, BUTR retains the same model size and adds 0.6--1.0 ms of T4 deployment latency relative to D-FINE, while achieving higher training efficiency and better overall AP across scales; for example, BUTR-L reaches 35.7 AP versus 35.2 for D-FINE-L, with about half as many queries on average (316 vs.\ 600) and slightly lower computation (326.1 vs.\ 329.6 GFLOPs).
Compared with dynamic-query detectors, BUTR is both more accurate and substantially more efficient; for example, BUTR achieves higher AP with fewer parameters and lower computation,
while requiring about $2.1\times$ less training time and delivering $1.6$--$1.9\times$ faster T4 inference than same-scale Dome-DETR.

\textbf{VisDrone.}
Compared with AI-TOD-v2, VisDrone has a milder density tail (max $\#\mathrm{GT}{=}461$, with only one image exceeding $400$ objects), so the gains from dynamic query are expected to be smaller under our budgeting view.
Tab.~\ref{tab:visdrone_results} supports this expectation.
In this regime, BUTR achieves higher accuracy than Dome-DETR with $1.5$--$1.8\times$ faster T4 inference. Relative to D-FINE, it achieves a slight AP gain with similar GFLOPs and 1.7--2.0 ms of additional deployment latency.

\begin{table}[htbp]
\centering
\fontsize{8.2}{8.8}\selectfont
\captionsetup{width=0.98\linewidth, font=small}
\setlength{\tabcolsep}{2.1pt}
\renewcommand{\arraystretch}{1.12}
\caption{Comparison on VisDrone. Results are from our re-implementations. Dynamic-query GFLOPs and $|Q|$ are averages. Tr. (h): training time (hours); Lat. (ms): T4 TensorRT latency (milliseconds).}
\vspace{2pt}
\label{tab:visdrone_results}
\begin{tabular}{l|ccccc|cccccc}
\toprule
\textbf{Method} & \textbf{\#Params} & \textbf{GFLOPs} & $\boldsymbol{|Q|}$ & \textbf{Tr. (h)} & \textbf{Lat. (ms)} & \textbf{AP} & \textbf{AP}$_{50}$ & \textbf{AP}$_{75}$ & \textbf{AP}$_{s}$ & \textbf{AP}$_m$ & \textbf{AP}$_l$ \\
\midrule

\rowcolor{orange!8}
\multicolumn{12}{c}{\textbf{Fixed-query detectors}} \\
D-FINE-S~\cite{peng2024d} & 11.6M & 113.3 & 400 & 6.7 & 11.39 & 26.6 & 45.1 & 27.0 & 16.8 & 37.0 & 53.6 \\
D-FINE-M~\cite{peng2024d} & 22.3M & 245.9 & 400 & 9.2 & 17.35 & 29.2 & 48.9 & 30.0 & 18.9 & 40.5 & 53.8 \\
D-FINE-L~\cite{peng2024d} & 34.5M & 329.5 & 400 & 11.9 & 22.06 & 29.6 & 49.6 & 30.2 & 19.0 & 40.8 & 55.7 \\
DEIM-S~\cite{huang2025deim} & 11.6M & 113.3 & 400 & 8.5 & 11.23 & 21.9 & 41.1 & 22.4 & 17.3 & 27.5 & 31.7 \\
DEIM-M~\cite{huang2025deim} & 22.3M & 245.9 & 400 & 10.7 & 17.40 & 24.5 & 44.6 & 24.9 & 18.4 & 31.8 & 36.8 \\
DEIM-L~\cite{huang2025deim} & 34.5M & 329.5 & 400 & 13.4 & 22.37 & 25.2 & 45.4 & 25.2 & 18.4 & 32.3 & 38.2 \\
\midrule

\rowcolor{orange!8}
\multicolumn{12}{c}{\textbf{Dynamic-query detectors}} \\
DQ-DETR~\cite{huang2024dq} & 58.7M & 1782.4 & 512 & 4.5 & 211.18 & 18.4 & 33.7 & 17.9 & 12.0 & 27.0 & 35.8 \\
Dome-DETR-S~\cite{hu2025dome} & 13.2M & 162.9 & 389 & 14.0 & 23.25 & 26.4 & 45.1 & 26.8 & 16.4 & 36.9 & 51.0 \\
Dome-DETR-M~\cite{hu2025dome} & 23.9M & 295.4 & 384 & 19.9 & 29.69 & 28.9 & 48.5 & 29.5 & 18.8 & 40.0 & 55.2 \\
Dome-DETR-L~\cite{hu2025dome} & 36.0M & 379.0 & 388 & 22.4 & 34.47 & 29.3 & 49.2 & 30.0 & 18.6 & 40.8 & 56.4 \\
\cmidrule(lr){1-12}
\textbf{BUTR-S} (Ours) & 11.6M & 114.9 & 416 & 6.8\hspace{1pt}\fontsize{7.8}{9}\selectfont(\textcolor{PosColor}{2.1$\times$}) & 13.26\hspace{1pt}\fontsize{7.8}{9}\selectfont(\textcolor{PosColor}{1.8$\times$}) & \textbf{26.7}\hspace{1pt}\textbf{\fontsize{7.8}{9}\selectfont(\textcolor{PosColor}{$+$0.3})} & 45.2 & 27.4 & 16.8 & 37.3 & 52.5 \\
\textbf{BUTR-M} (Ours) & 22.3M & 245.8 & 396 & 9.3\hspace{1pt}\fontsize{7.8}{9}\selectfont(\textcolor{PosColor}{2.1$\times$}) & 19.31\hspace{1pt}\fontsize{7.8}{9}\selectfont(\textcolor{PosColor}{1.5$\times$}) & \textbf{29.4}\hspace{1pt}\textbf{\fontsize{7.8}{9}\selectfont(\textcolor{PosColor}{$+$0.5})} & 49.4 & 29.9 & 18.9 & 40.7 & 57.0 \\
\textbf{BUTR-L} (Ours) & 34.5M & 329.5 & 398 & 11.8\hspace{1pt}\fontsize{7.8}{9}\selectfont(\textcolor{PosColor}{1.9$\times$}) & 23.76\hspace{1pt}\fontsize{7.8}{9}\selectfont(\textcolor{PosColor}{1.5$\times$}) & \textbf{29.8}\hspace{1pt}\textbf{\fontsize{7.8}{9}\selectfont(\textcolor{PosColor}{$+$0.5})} & 50.1 & 30.4 & 19.1 & 41.1 & 57.1 \\
\bottomrule
\end{tabular}
\end{table}

\begin{figure}[h]
\centering
\captionsetup{width=0.98\linewidth, font=small}
\begin{subfigure}[t]{0.47\linewidth}
    \centering
    \includegraphics[width=\linewidth]{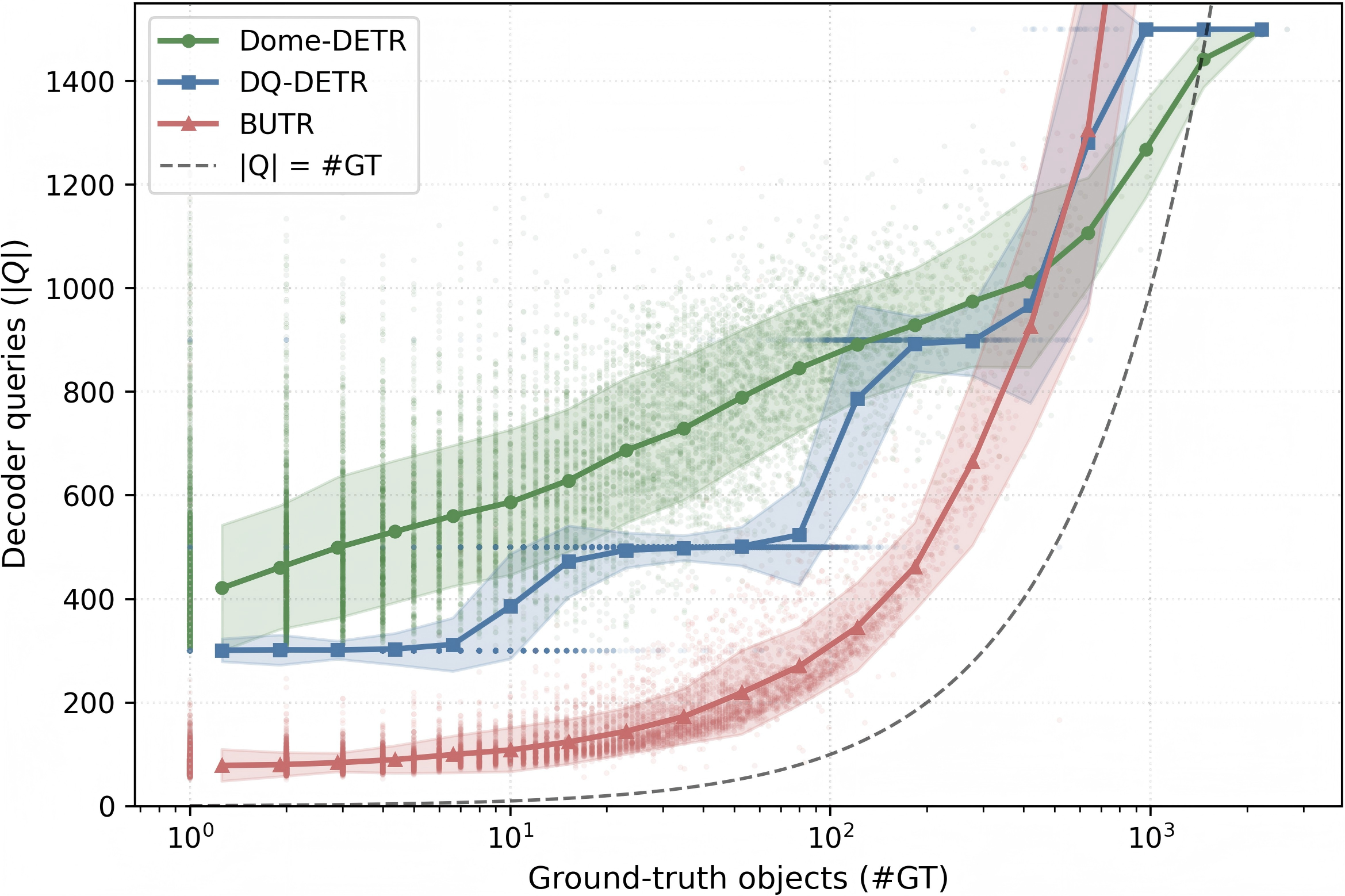}
    \caption{AI-TOD-v2}
    \label{fig:aitodv2_map_vs_k_stratified}
\end{subfigure}
\hfill
\begin{subfigure}[t]{0.47\linewidth}
    \centering
    \includegraphics[width=\linewidth]{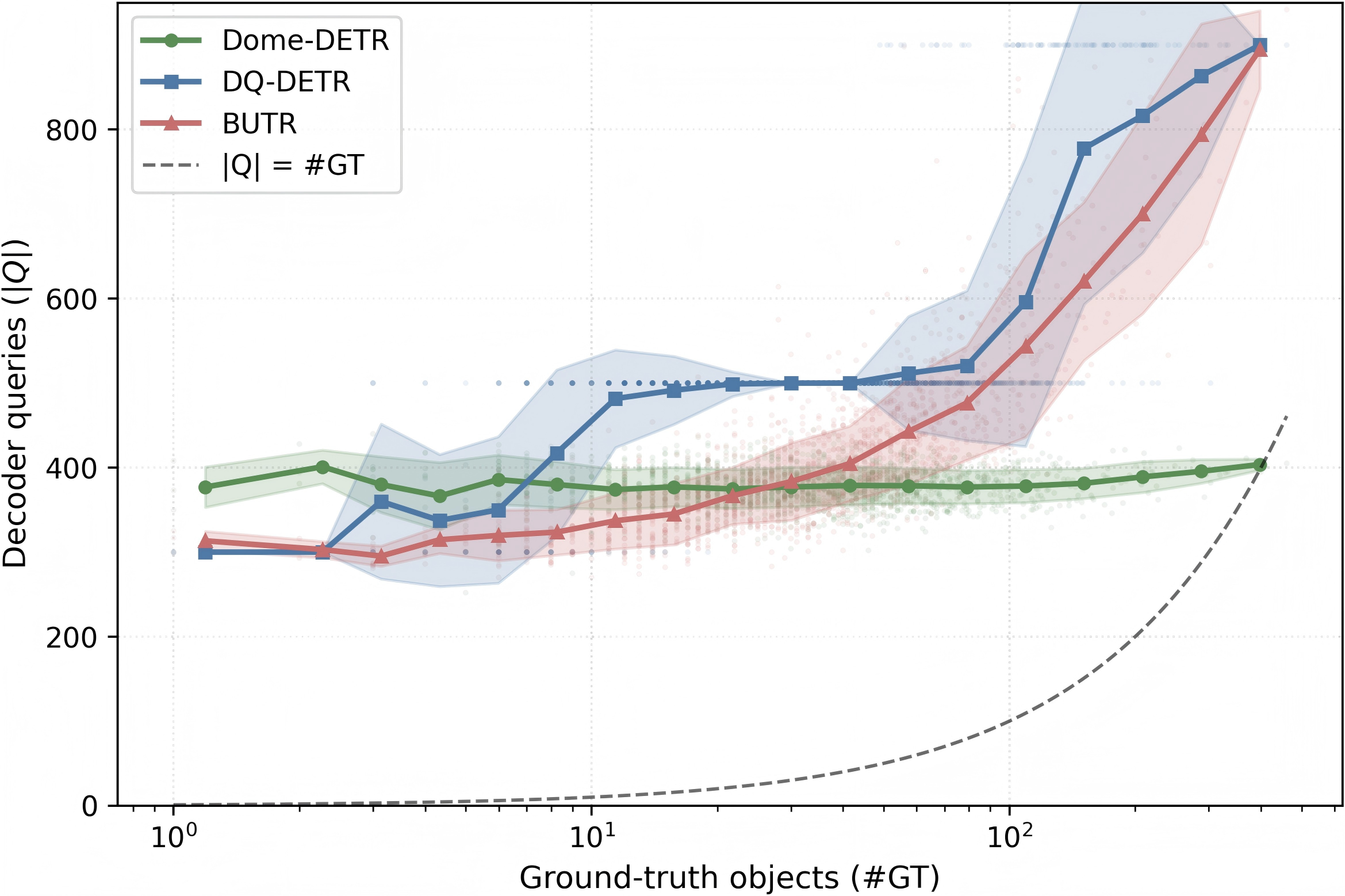}
    \caption{VisDrone}
    \label{fig:visdrone_map_vs_k_stratified}
\end{subfigure}
\caption{Query usage versus image density.
Each point corresponds to one test image; solid lines and shaded regions denote the binned mean and standard deviation, respectively.
The dashed gray line is an ideal reference proportional to \#GT.
BUTR shows the clearest monotonic increase with image density.}
\label{fig:budget_vs_gt}
\end{figure}

\textbf{Query usage behavior.}
Fig.~\ref{fig:budget_vs_gt} provides a direct view of the budgeting behavior of dynamic-query methods. We disable the dataset-specific clipping to observe BUTR's native budget estimates. BUTR exhibits a clear increasing trend in allocated queries with \#GT, whereas DQ-DETR is limited by discrete budget bins and Dome-DETR shows weaker alignment with density.

\subsection{Density-Stratified Evaluation}
\label{sec:density_stratified}
\begin{figure}[htbp]
\centering
\captionsetup{width=0.98\linewidth, font=small}
\begin{subfigure}[t]{0.47\linewidth}
    \centering
    \includegraphics[width=\linewidth]{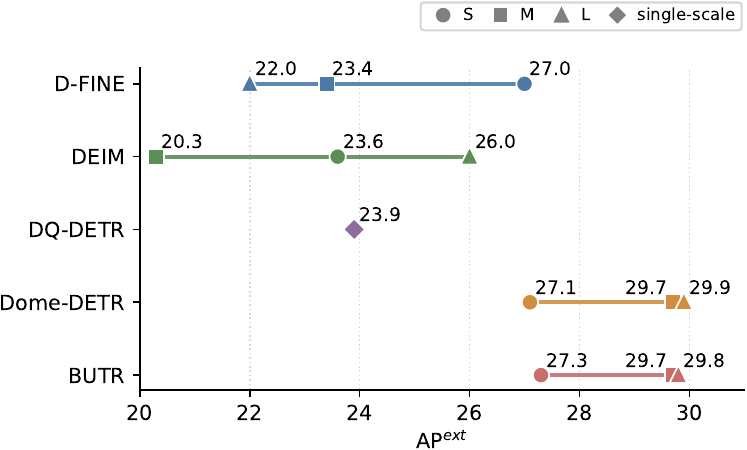}
    \caption{AI-TOD-v2: \texttt{extreme} subset}
    \label{fig:aitodv2_extreme_dotplot}
\end{subfigure}
\hfill
\begin{subfigure}[t]{0.47\linewidth}
    \centering
    \includegraphics[width=\linewidth]{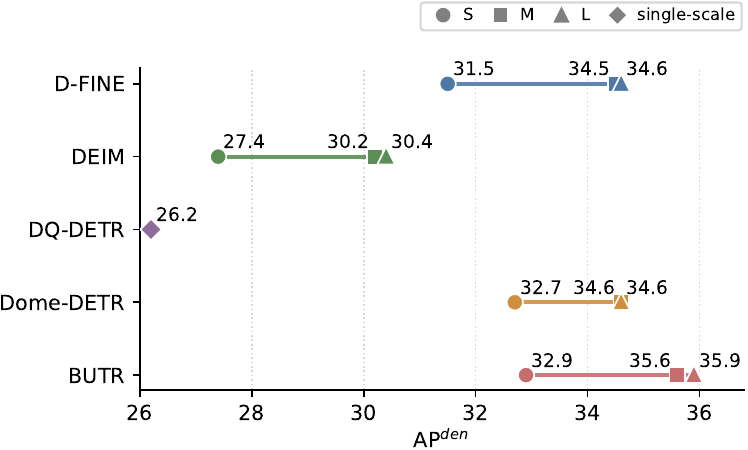}
    \caption{VisDrone: \texttt{dense} subset}
    \label{fig:visdrone_dense_dotplot}
\end{subfigure}
\caption{Density-stratified tail performance across model families and scales. (a) AP$^{ext}$ on the \texttt{extreme} subset of AI-TOD-v2. (b) AP$^{den}$ on the \texttt{dense} subset of VisDrone.}
\label{fig:density_stratified_figure}
\end{figure}

Overall AP can obscure performance on the rare high-density tail.
Following Sec.~\ref{sec:diagnosis}, we examine the \texttt{extreme} subset of AI-TOD-v2 (AP$^{ext}$) and the \texttt{dense} subset of VisDrone (AP$^{den}$), where insufficient decoder capacity is particularly consequential.
Fig.~\ref{fig:aitodv2_extreme_dotplot} shows that the benefits of dynamic query on AI-TOD-v2 are more pronounced on the tail subset than in overall AP.
Increasing model scale, however, does not consistently improve tail performance for fixed-query detectors: D-FINE achieves 27.0, 23.4, and 22.0 AP$^{ext}$ at the S, M, and L scales, respectively.
BUTR remains competitive with Dome-DETR on AP$^{ext}$ and achieves higher AP$^{den}$ across scales (Fig.~\ref{fig:visdrone_dense_dotplot}).
Its gains also extend beyond the tail, as the full results in App.~\ref{app:density_stratified} show improvements across multiple density strata.

Because the tail subsets contain few images, we examine AP$^{ext}$ stability for the L-scale models on AI-TOD-v2.
Under both repeated seed-0 runs and runs with matched seeds 0, 1, and 2, BUTR-L exhibits lower AP$^{ext}$ variability than Dome-DETR-L.
App.~\ref{app:training_stability} provides the protocols and results.


\subsection{Ablations}
\label{sec:exp_ablation}
Tab.~\ref{tab:butr_ablation} ablates two aspects of BUTR-L on AI-TOD-v2: the estimator $e$ and the mapping $g$.
For the estimator, all variants are retrained and adopt our budget rule (Eq.~\ref{eq:budget_rule}) with $\gamma=1.0$.
Specifically, we compare four instantiations of the ranking function $\phi$ that defines $s_i = \phi(\bm{\ell}_i)$:
\textit{Max} with $s_i=\max(0,\max_c \bm{\ell}_{ic})$,
\textit{Max-sigmoid} with $s_i=\max_c \sigma(\bm{\ell}_{ic})$,
\textit{Sum-prob} with $s_i=\sum_c \sigma(\bm{\ell}_{ic})$, and
\textit{Union-prob} with $s_i=1-\prod_c(1-\sigma(\bm{\ell}_{ic}))$.
The results show that BUTR is not sensitive to the exact choice of $\phi$.
Among them, we choose the \textit{Max-sigmoid} proxy as the default setting.
We then ablate the budget mapping using \textit{Max-sigmoid} estimator, considering two variants: (i) varying the safety factor $\gamma$ in Eq.~\ref{eq:budget_rule}, and (ii) a discrete binning strategy that quantizes the soft density proxy $e(x)$ into one of a set of predefined budget levels. Specifically, we adopt the budget bin setting $\{300,500,900,1500\}$ from DQ-DETR~\cite{huang2024dq} as the discrete targets.
We find that the model is robust to $\gamma$ over a wide range (1.0--3.0). The discrete binning baseline achieves comparable accuracy but requires manually designed thresholds. We adopt $\gamma=2.0$ as the default setting.

\begin{table}[htbp]
\centering
\fontsize{8.2}{8.6}\selectfont
\captionsetup{width=0.98\linewidth, font=small}
\setlength{\tabcolsep}{3pt}
\renewcommand{\arraystretch}{1.10}
\caption{Ablations of budget estimation and mapping methods for BUTR-L on AI-TOD-v2.
Estimator $e$ variants are trained and evaluated without an additional mapping (i.e., $\gamma=1.0$), while the mapping $g$ study is conducted by inference-time ablation using the \textit{Max-sigmoid} estimator. Tr. (h): training wall-clock time.}
\vspace{2pt}
\label{tab:butr_ablation}
\begin{tabular}{l|l|cc|cccccccc}
\toprule
\textbf{Category} & \textbf{Variant} & $\boldsymbol{|Q|}$ & \textbf{Tr. (h)} & \textbf{AP} & \textbf{AP}$_{50}$ & \textbf{AP}$_{75}$ & \textbf{AP}$_{vt}$ & \textbf{AP}$_t$ & \textbf{AP}$_s$ & \textbf{AP}$_m$ & \textbf{AP}$^{ext}$ \\
\midrule
\multirow{4}{*}{\makecell{Estimator $e$}}
& \textit{Max} & 302 & 37.8 & 35.6 & 69.6 & 31.5 & 20.4 & 36.4 & 40.4 & 49.0 & 27.7 \\
& \textit{Max-sigmoid} & 303 & 37.8 & 35.5 & 69.3 & 32.2 & 21.1 & 36.1 & 40.5 & 49.1 & 28.9 \\
& \textit{Sum-prob} & 304 & 41.5 & 35.3 & 69.4 & 31.1 & 20.3 & 36.0 & 39.6 & 47.0 & 31.0 \\
& \textit{Union-prob} & 304 & 40.7 & 35.2 & 68.9 & 31.4 & 18.9 & 36.0 & 40.2 & 48.5 & 28.3 \\
\midrule
\multirow{5}{*}{Mapping $g$}
& $\gamma=1.2$ & 305 & / & 35.6 & 69.4 & 32.2 & 21.2 & 36.2 & 40.6 & 49.1 & 29.5 \\
& $\gamma=1.5$ & 308 & / & 35.6 & 69.5 & 32.2 & 21.2 & 36.2 & 40.6 & 49.2 & 29.6 \\
& \textbf{$\gamma=2.0$} & 316 & / & 35.7 & 69.6 & 32.3 & 21.2 & 36.2 & 40.7 & 49.2 & 29.8 \\
& $\gamma=3.0$ & 340 & / & 35.7 & 69.6 & 32.3 & 21.2 & 36.3 & 40.7 & 49.3 & 29.8 \\
& binning & 305 & / & 35.6 & 69.6 & 32.2 & 21.2 & 36.2 & 40.6 & 49.2 & 29.5 \\
\bottomrule
\end{tabular}
\end{table}

\subsection{Generality beyond Tiny Object Scenarios}
\label{sec:generality}
We first extend the evaluation to dense agricultural imagery using Sorghum-Grain-Count~\cite{bakshi2025sorghum}, training all compared models on the cropped agricultural dataset as in our AI-TOD-v2 and VisDrone evaluations.
BUTR-L matches or exceeds the overall AP of fixed-query baselines using 1100 queries while decoding only 378 queries on average; the full protocol and results are provided in App.~\ref{app:sorghum_protocol}.

We then evaluate BUTR on COCO~\cite{lin2014microsoft} using D-FINE~\cite{peng2024d} and DEIM~\cite{huang2025deim} under an \emph{inference-time plug-in} protocol.
Unlike the preceding evaluations, we keep the officially provided checkpoints fixed and replace only the decoder query initialization with our budgeter, without retraining.
This paired comparison isolates the effect of query budgeting from training-run variability, allowing small AP differences to be examined with identical model weights.
As shown in Tab.~\ref{tab:generality}, BUTR reduces the average number of decoder queries and GFLOPs across all variants while maintaining or slightly improving overall AP (by 0.1--0.2 points).
We also report paired T4 TensorRT latency for the fixed-query and BUTR-augmented variants under the protocol in App.~\ref{app:t4_deployment}.

\begin{table}[htbp]
\centering
\fontsize{8.2}{8.6}\selectfont
\captionsetup{width=0.98\linewidth, font=small}
\setlength{\tabcolsep}{2pt}
\renewcommand{\arraystretch}{1.12}
\caption{Generality of BUTR on COCO under an inference-time plug-in setting without retraining. We report average GFLOPs and $|Q|$ for the BUTR-augmented variants. Lat. (ms): T4 TensorRT latency (App.~\ref{app:t4_deployment}).}
\vspace{2pt}
\label{tab:generality}
\begin{tabular}{l|cccc|cccccc}
\toprule
\textbf{Method} & \textbf{\#Params} & \textbf{GFLOPs} & $\boldsymbol{|Q|}$ & \textbf{Lat. (ms)} & \textbf{AP}$^{val}$ & \textbf{AP}$^{val}_{50}$ & \textbf{AP}$^{val}_{75}$ & \textbf{AP}$^{val}_{s}$ & \textbf{AP}$^{val}_{m}$ & \textbf{AP}$^{val}_{l}$ \\
\midrule
D-FINE-S~\cite{peng2024d} & 10.2M & 25.2 & 300 & 3.63 & 48.5 & 65.4 & 52.6 & 29.3 & 52.2 & 65.4 \\
\rowcolor{gray!10}\textbf{D-FINE-S+BUTR} & 10.2M & 24.6 & 232 & 4.24 & \textbf{48.6}\hspace{1pt}\textbf{\fontsize{7.8}{9}\selectfont(\textcolor{PosColor}{$+$0.1})} & 65.5 & 52.8 & 29.3 & 52.1 & 65.2 \\

D-FINE-M~\cite{peng2024d} & 19.3M & 56.7 & 300 & 5.66 & 52.3 & 69.9 & 56.4 & 33.2 & 56.5 & 70.2 \\
\rowcolor{gray!10}\textbf{D-FINE-M+BUTR} & 19.3M & 56.6 & 289 & 6.44 & \textbf{52.4}\hspace{1pt}\textbf{\fontsize{7.8}{9}\selectfont(\textcolor{PosColor}{$+$0.1})} & 70.0 & 56.5 & 33.2 & 56.6 & 70.1 \\

D-FINE-L~\cite{peng2024d} & 30.8M & 91.0 & 300 & 7.94 & 54.0 & 71.6 & 58.4 & 36.5 & 58.0 & 71.9 \\
\rowcolor{gray!10}\textbf{D-FINE-L+BUTR} & 30.8M & 90.2 & 253 & 8.81 & \textbf{54.1}\hspace{1pt}\textbf{\fontsize{7.8}{9}\selectfont(\textcolor{PosColor}{$+$0.1})} & 71.8 & 58.4 & 36.6 & 58.0 & 72.0 \\

D-FINE-X~\cite{peng2024d} & 61.6M & 202.5 & 300 & 13.52 & 55.8 & 73.7 & 60.2 & 37.3 & 60.5 & 73.5 \\
\rowcolor{gray!10}\textbf{D-FINE-X+BUTR} & 61.6M & 201.7 & 253 & 13.84 & \textbf{55.9}\hspace{1pt}\textbf{\fontsize{7.8}{9}\selectfont(\textcolor{PosColor}{$+$0.1})} & 73.9 & 60.4 & 37.4 & 60.6 & 73.4 \\

\cmidrule(lr){1-11}
DEIM-S~\cite{huang2025deim} & 10.2M & 25.2 & 300 & 3.75 & \textbf{49.0} & 65.9 & 53.1 & 30.4 & 52.6 & 65.7 \\
\rowcolor{gray!10}\textbf{DEIM-S+BUTR} & 10.2M & 24.6 & 239 & 4.32 & \textbf{49.0}\hspace{1pt}\textbf{\fontsize{7.8}{9}\selectfont(\textcolor{NeuColor}{$\pm$0.0})} & 66.0 & 53.2 & 30.2 & 52.6 & 65.5 \\

DEIM-M~\cite{huang2025deim} & 19.3M & 56.7 & 300 & 5.73 & 52.7 & 70.0 & 57.3 & 35.2 & 56.7 & 69.4 \\
\rowcolor{gray!10}\textbf{DEIM-M+BUTR} & 19.3M & 55.9 & 236 & 6.38 & \textbf{52.8}\hspace{1pt}\textbf{\fontsize{7.8}{9}\selectfont(\textcolor{PosColor}{$+$0.1})} & 70.1 & 57.4 & 35.1 & 56.7 & 69.5 \\

DEIM-L~\cite{huang2025deim} & 30.8M & 91.1 & 300 & 8.07 & 54.7 & 72.4 & 59.4 & 36.9 & 59.6 & 71.8 \\
\rowcolor{gray!10}\textbf{DEIM-L+BUTR} & 30.8M & 89.8 & 224 & 8.83 & \textbf{54.9}\hspace{1pt}\textbf{\fontsize{7.8}{9}\selectfont(\textcolor{PosColor}{$+$0.2})} & 72.6 & 59.6 & 37.1 & 59.8 & 71.8 \\

DEIM-X~\cite{huang2025deim} & 61.7M & 202.7 & 300 & 13.78 & 56.5 & 74.0 & 61.5 & 38.8 & 61.4 & 74.2 \\
\rowcolor{gray!10}\textbf{DEIM-X+BUTR} & 61.7M & 200.6 & 183 & 13.71 & \textbf{56.6}\hspace{1pt}\textbf{\fontsize{7.8}{9}\selectfont(\textcolor{PosColor}{$+$0.1})} & 74.2 & 61.7 & 38.9 & 61.4 & 74.0 \\
\bottomrule
\end{tabular}
\end{table}

Taken together, BUTR improves accuracy and efficiency on AI-TOD-v2, which exhibits the heavier density imbalance and more extreme high-density tail of the two aerial benchmarks; its benefits remain evident on the milder VisDrone benchmark and extend to dense agricultural imagery; and even on conventional COCO, it reduces decoder computation without sacrificing overall AP or requiring retraining.
These results highlight two complementary advantages of BUTR: compared with existing dynamic-query designs, it avoids the overhead and potential decision errors introduced by additional learned predictors and heuristic filtering in query initialization; compared with fixed-query detectors, it adapts decoder capacity to address the risks of insufficient recall in crowded images under a small budget and increased false positives in sparse images during deployment under a large budget.
This combination supports efficient and effective detection across the four datasets evaluated in this work.
More broadly, because BUTR builds on proposal confidences and confidence-ranked initialization already present in two-stage detectors, we expect its budgeting approach to be applicable to a broader range of detection scenarios.

\section{Conclusion}
\label{sec:conclusion}

We revisited dynamic query for tiny object detection from a mechanism-first perspective, finding that adaptive budgeting largely explains Dome-DETR's accuracy in the evaluated settings.
Motivated by this view, we proposed BUTR, a principled dynamic query framework that estimates a per-image decoder budget from encoder confidences while retaining standard Top-$K$ initialization form.
Experiments on AI-TOD-v2 and VisDrone show improved overall accuracy and efficiency over Dome-DETR, while inference-only integration on COCO reduces queries and computation without degrading overall AP.
We hope this perspective informs the future design and evaluation of detectors.

\textbf{Limitations.}
BUTR relies on encoder confidences as its budgeting signal. Unlike fixed Top-$K$ selection, which depends on score ranking, score aggregation also depends on their numerical scale; changes in score calibration, proposal count, or domain may therefore affect allocation even when rankings remain unchanged.
Our experiments support applicability in the tested domains, while other dense settings, such as crowded products and cell or particle imagery, remain to be evaluated.
Dynamic-budget inference also incurs synchronization and engine-boundary overhead in our current TensorRT deployment, so fewer queries do not necessarily translate into lower latency.
Finally, the high-density subsets contain few images (26 on AI-TOD-v2 and 7 on VisDrone), so their AP is sensitive to test-sample composition. Density-stratified evaluation should therefore complement, rather than replace, overall evaluation.

\section*{Acknowledgments}
This work was supported by the New Generation Artificial Intelligence-National Science and Technology Major Project (Nos. 2025ZD0123502),  National Natural Science Fund of China (Nos. 62506155), Provincial Natural Science Fund of Jiangsu (Nos. BK20251985), Suzhou Municipal Leading Talents Fund (Nos. ZXL2025320), ZTE Industry-University Project (Nos. IA20260910004), Alipay Industry-University Project (Nos. 2026101543758), and Key Project of the State Key Laboratory for Novel Software Technology (Nos. ZZKT2026A14).

\bibliographystyle{unsrt}
\bibliography{main}

\clearpage
\appendix
\makeatletter
\setlength{\@fptop}{0pt}
\makeatother

\section{Supplementary Comparison with DDQ-Style Initialization}
\label{app:ddq_style}

\textbf{Why examine DDQ-style initialization?}
The rationale for choosing DDQ is straightforward: if our PAQI-derived conclusion holds, replacing confidence-ranked Top-$K$ with an initializer that changes \emph{which} queries are retained while keeping \emph{how many} queries fixed should yield nearly unchanged performance.
DDQ~\cite{zhang2023dense} provides such a mechanism: its distinct-query selection takes a dense set of query candidates and retains a fixed-size, spatially distinct subset.
For each dataset, we separately train and evaluate models with DDQ-style initialization and the original Top-$K$ initialization at the same query count, using each initializer during both training and inference.
We emphasize that the full DDQ framework also includes inter-layer query selection and corresponding loss designs; here, we adopt only its distinct query initialization mechanism.

\textbf{Why use D-FINE?}
Removing PAQI and its associated components from Dome-DETR leaves an architecture close to D-FINE~\cite{peng2024d}.
D-FINE therefore provides a clean base that directly connects this comparison to our PAQI diagnosis: we replace its original Top-$K$ initializer with DDQ-style selection while keeping the query count fixed.

\textbf{Result and interpretation.}
As shown in Tab.~\ref{tab:ddq_style}, the separately trained variants achieve the same overall AP: 35.2 on AI-TOD-v2 and 29.6 on VisDrone, using fixed budgets of 600 and 400 queries, respectively.
Thus, even when the detector is trained with the alternative initializer, DDQ-style selection yields no observed overall AP advantage over the original initializer in this comparison.
This complements the PAQI diagnosis: standard confidence-ranked initialization remains competitive without the additional selection strategy, supporting our decision to focus BUTR on budget allocation.

\begin{table}[htbp]
\centering
\captionsetup{width=0.98\linewidth, font=small}
\fontsize{8.2}{8.8}\selectfont
\caption{DDQ-style initialization within D-FINE-L. Each variant is separately trained with its initializer, which is also used at inference. Original Top-$K$ results are those reported in Tabs.~\ref{tab:aitodv2_results} and~\ref{tab:visdrone_results}.}
\vspace{2pt}
\label{tab:ddq_style}
\begin{tabular}{lcccc}
\toprule
\multirow{2}{*}{\textbf{Initialization}} & \multicolumn{2}{c}{\textbf{AI-TOD-v2}} & \multicolumn{2}{c}{\textbf{VisDrone}} \\
\cmidrule(lr){2-3}\cmidrule(lr){4-5}
 & $|Q|$ & AP & $|Q|$ & AP \\
\midrule
Original Top-$K$ & 600 & 35.2 & 400 & 29.6 \\
DDQ-style & 600 & 35.2 & 400 & 29.6 \\
\bottomrule
\end{tabular}
\end{table}

\section{Algorithmic Description of BUTR}
\label{app:butr_algorithm}

Alg.~\ref{alg:butr} summarizes the BUTR query initialization module inserted after the encoder proposal generation stage of a two-stage DETR.
Given the encoder proposals from the host detector, BUTR computes proposal-level ranking scores from classification logits and aggregates them into an image-level density proxy.
This proxy is mapped to a decoder query budget through a safety factor; in practice, the budget can be optionally clipped to a feasible range imposed by the detector or comparison protocol.
The resulting budget is then used in the standard Top-$K(x)$ initialization form.
Therefore, BUTR modifies only the decoder query budget using existing encoder outputs and proposal confidences, without adding new learned modules or changing the backbone, encoder, decoder, or detection heads.

\begin{algorithm}[h]
\caption{BUTR Query Initialization}
\label{alg:butr}
\begin{algorithmic}[1]
\Require Encoder proposals $P=\{(\bm{z}_i,\bm{r}_i,\bm{\ell}_i)\}_{i=1}^{L}$ from a two-stage DETR.
\Require Safety factor $\gamma\ge1$; optional feasible range $1\le K_{\min}\le K_{\max}\le L$, with $L\ge1$.
\Ensure Decoder query set $Q$.

\Statex \textit{// Proposal scoring}
\For{$i=1$ to $L$}
    \State Compute the proposal ranking score:
    $s_i \leftarrow \phi(\bm{\ell}_i) = \max_{c\in\{1,\dots,C\}} \sigma(\ell_{ic}).$
\EndFor

\Statex \textit{// Density proxy estimation}
\State Aggregate proposal scores as the image-level density proxy:
$e(x) \leftarrow \sum_{i=1}^{L} s_i.$

\Statex \textit{// Budget mapping}
\State Estimate the decoder query budget:
$\widetilde{K}(x) \leftarrow \lceil \gamma e(x) \rceil.$
\If{a feasible range $[K_{\min}, K_{\max}]$ is specified}
    \State
    $K(x) \leftarrow \operatorname{clip}\bigl(\widetilde{K}(x), K_{\min}, K_{\max}\bigr).$
\Else
    \State
    $K(x) \leftarrow \min\{L,\max\{1,\widetilde{K}(x)\}\}.$
\EndIf

\Statex \textit{// Standard Top-$K$ query initialization form}
\State Select the indices of the top $K(x)$ proposals according to $\{s_i\}_{i=1}^{L}$:
$\mathcal{I} \leftarrow \operatorname{TopK}(\{s_i\}_{i=1}^{L}, K(x)).$
\State Construct decoder queries from the selected proposals:
$Q \leftarrow \{(\bm{z}_i,\bm{r}_i)\mid i\in\mathcal{I}\}.$

\State \Return $Q$.
\end{algorithmic}
\end{algorithm}

\begin{figure}[b]
\centering
\captionsetup{width=0.98\linewidth, font=small}
\begin{subfigure}[t]{0.33\linewidth}
    \centering
    \includegraphics[width=\linewidth]{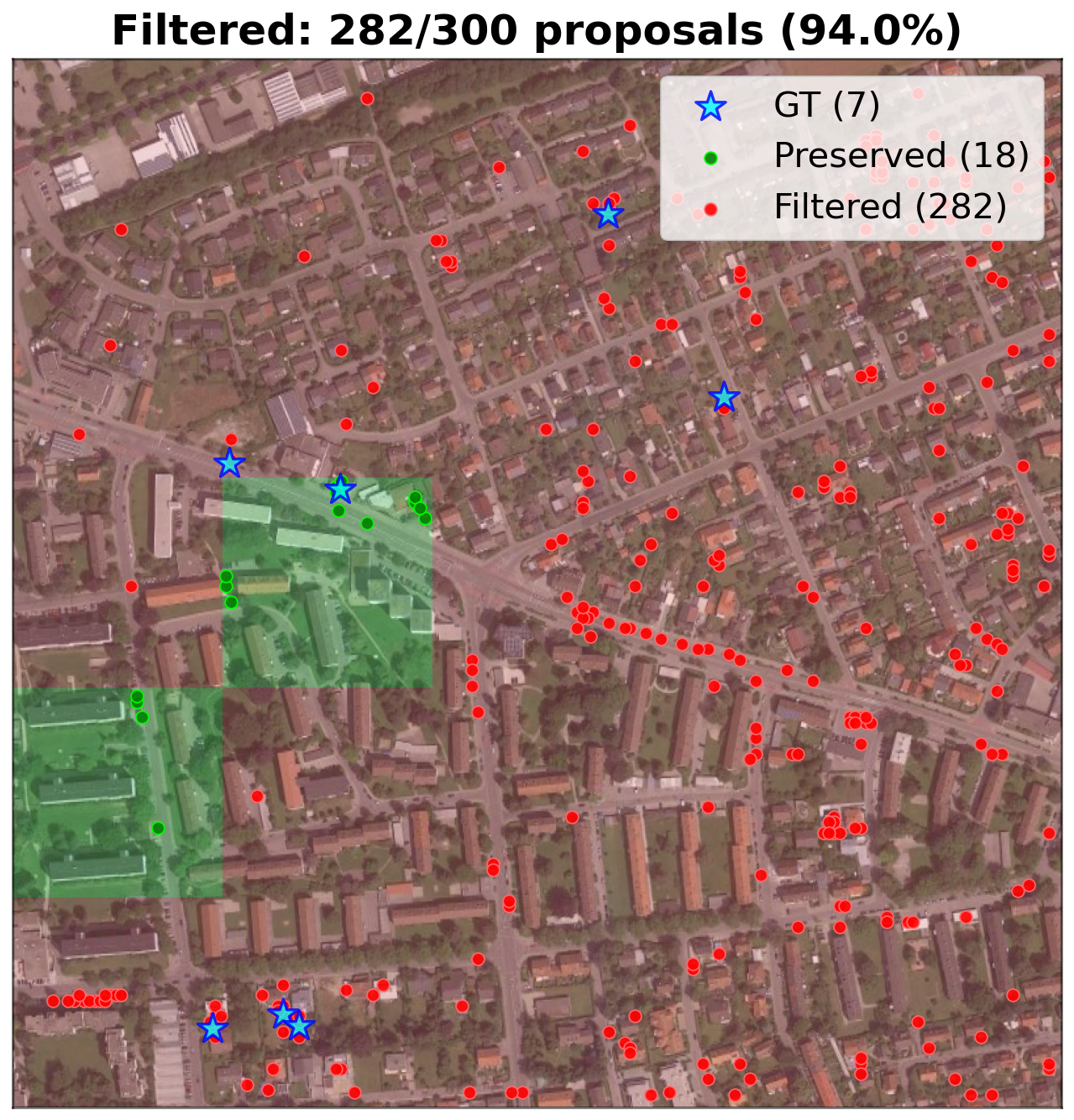}
    \label{fig:mask_filtering_0006}
\end{subfigure}\hfill
\begin{subfigure}[t]{0.33\linewidth}
    \centering
    \includegraphics[width=\linewidth]{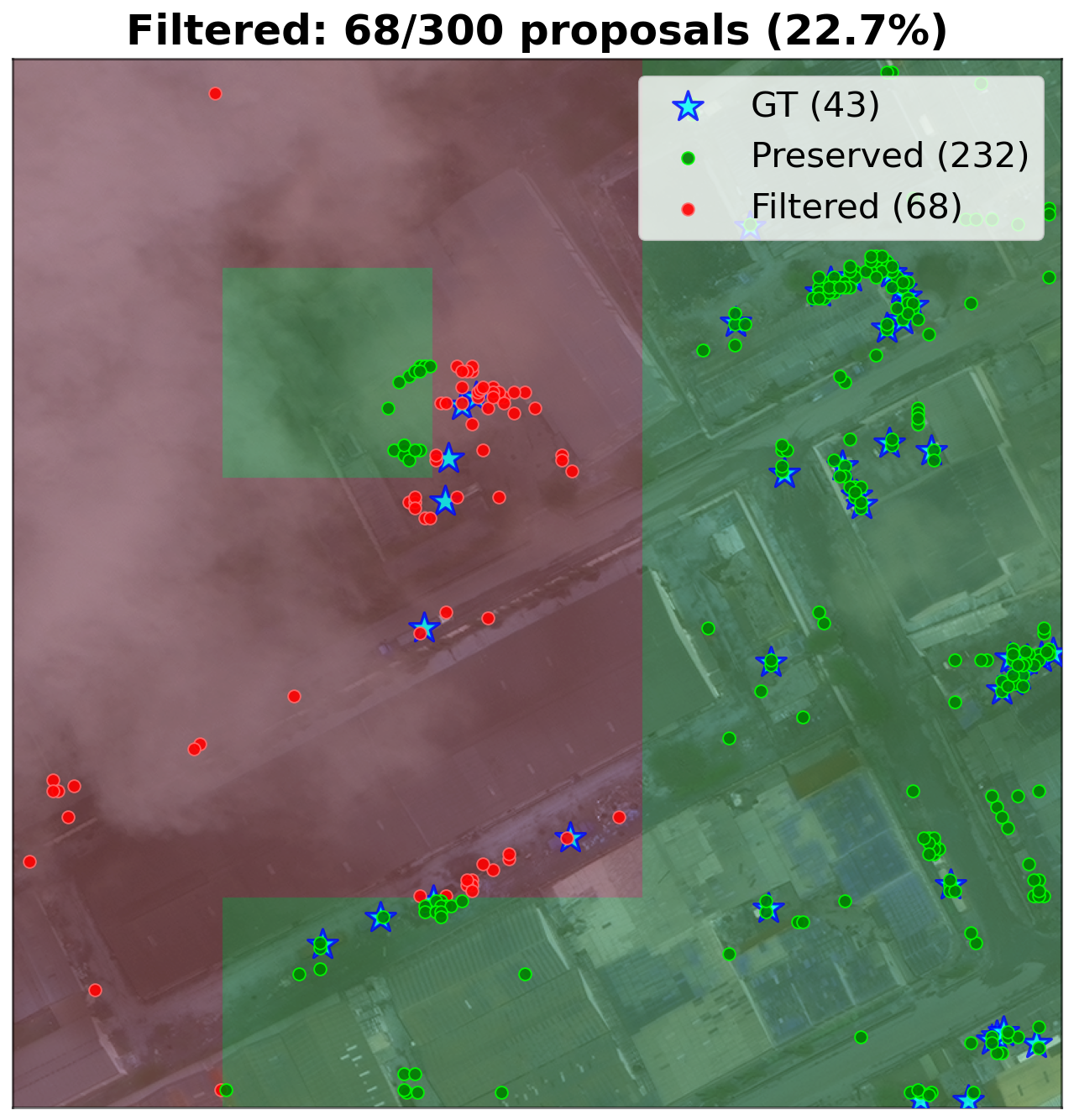}
    \label{fig:mask_filtering_0005}
\end{subfigure}\hfill
\begin{subfigure}[t]{0.33\linewidth}
    \centering
    \includegraphics[width=\linewidth]{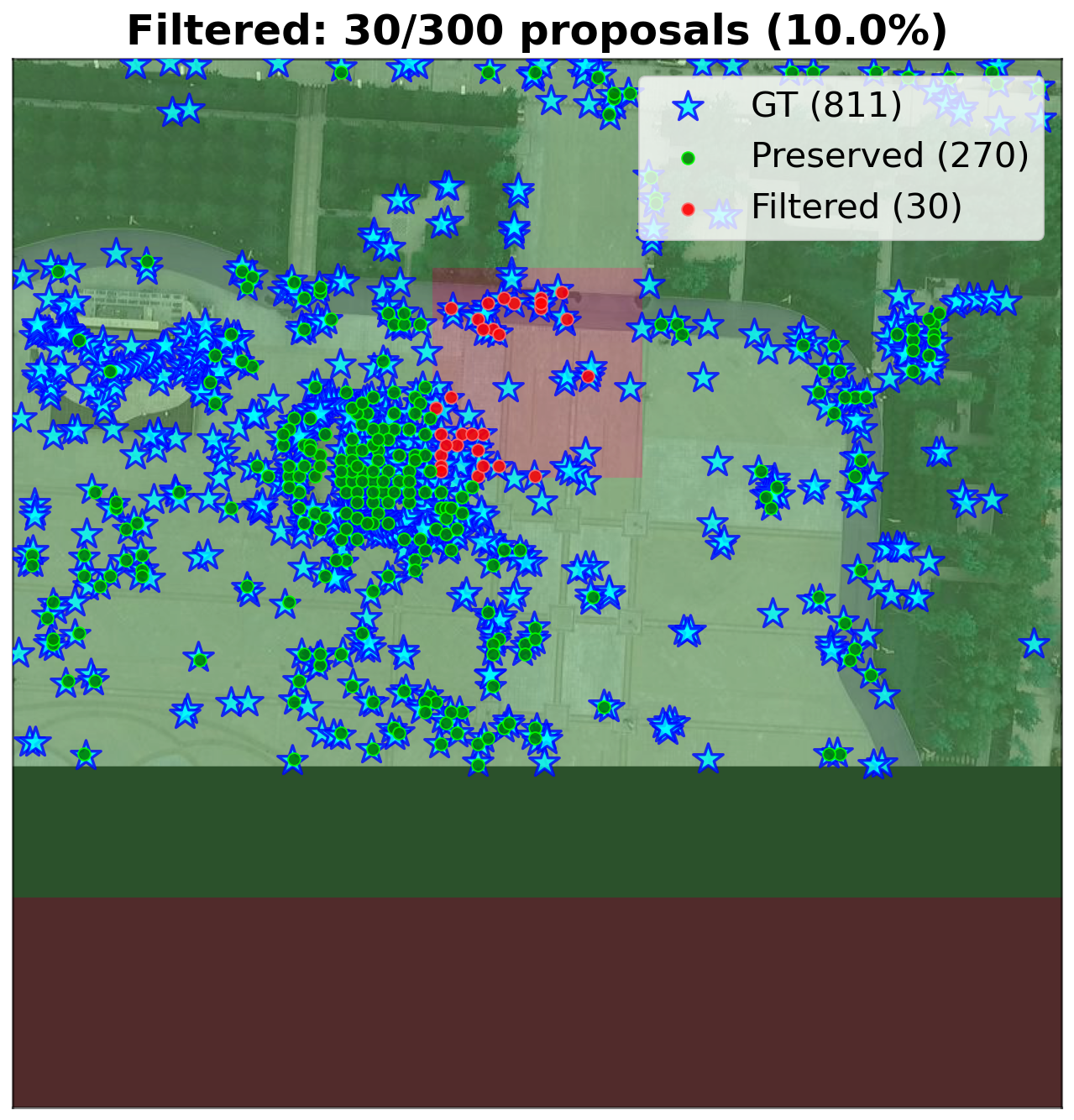}
    \label{fig:mask_filtering_0337}
\end{subfigure}
\vspace{-12pt}
\caption{Qualitative examples of window-mask filtering.
Green and red regions indicate mask-positive (preserved) and mask-negative (filtered) areas, respectively.
Cyan stars denote GT centers; green/red points are preserved/filtered high-confidence proposals.
Without Split, high-confidence proposals aligned with GT can be erroneously filtered.}
\label{fig:mask_filtering_examples}
\end{figure}

\section{Qualitative Examples of Erroneous Filtering}
\label{app:mask_filtering}
Fig.~\ref{fig:mask_filtering_examples} shows qualitative examples of the failure mode discussed in Sec.~\ref{sec:diagnosis}.
Without \op{Split}, the subsequent selection operators may filter out high-confidence proposals that are well aligned with GT objects.
As shown in the figure, some proposals near GT centers fall into mask-negative regions and are erroneously discarded, which helps explain the large AP drops in Rows~4--6 of Tab.~\ref{tab:paqi_ablation}.
These examples highlight the actual role of \op{Split}: it protects top-ranked proposals from harmful filtering.

\section{Why Budget-Matched Top-$K$ Reproduces PAQI}
\label{app:topk_overlap}
Two observations help explain the similar overall AP between PAQI and a Top-$K$ initializer with an approximately matched average budget: overlap between the selected queries and redundancy in encoder proposals. The overlap analysis below matches the query count per image, whereas the AP comparison in Tab.~\ref{tab:paqi_ablation} uses fixed budgets close to PAQI's average count.

\textbf{Overlap.}
We first examine whether PAQI selects nearly the same queries as a confidence-ranked Top-$|Q|$ initializer.
To this end, we define the overlap ratio as
\[
r_{\text{overlap}}
=
\frac{|Q_{\text{PAQI}} \cap Q_{\text{Top-}|Q|}|}{|Q_{\text{PAQI}}|},
\]
where $Q_{\text{PAQI}}$ denotes the queries selected by PAQI and $Q_{\text{Top-}|Q|}$ denotes the confidence-ranked Top-$|Q|$ proposals from the same encoder outputs.
The mean overlap is $71.5\%\pm 8.1\%$ on AI-TOD-v2 and $84.6\%\pm 1.9\%$ on VisDrone.
For reference, the random baseline is $\mathbb{E}[|Q|/K_{\max}]$, with standard deviation induced by per-image variation in $|Q|$, yielding $42.0\%\pm 15.3\%$ on AI-TOD-v2 and $77.7\%\pm 4.2\%$ on VisDrone.
These results show that PAQI is not identical to Top-$|Q|$, but already overlaps substantially with the confidence-ranked proposals, especially on VisDrone.

\begin{figure}[htbp]
\centering
\captionsetup{width=0.98\linewidth, font=small}
\begin{subfigure}[t]{0.495\linewidth}
    \centering
    \includegraphics[width=\linewidth]{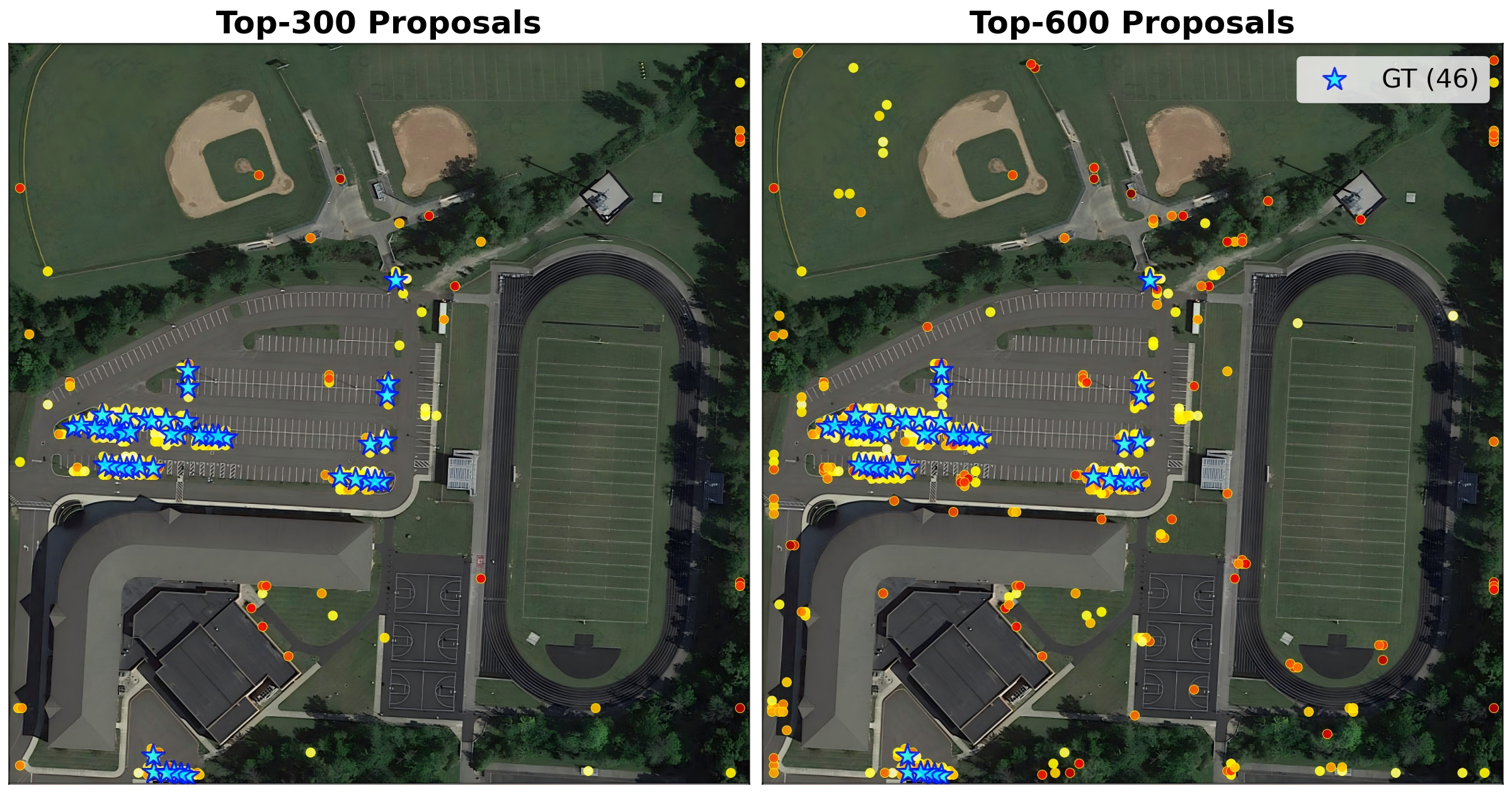}
    \label{fig:proposal_comparison_0094}
\end{subfigure}
\begin{subfigure}[t]{0.495\linewidth}
    \centering
    \includegraphics[width=\linewidth]{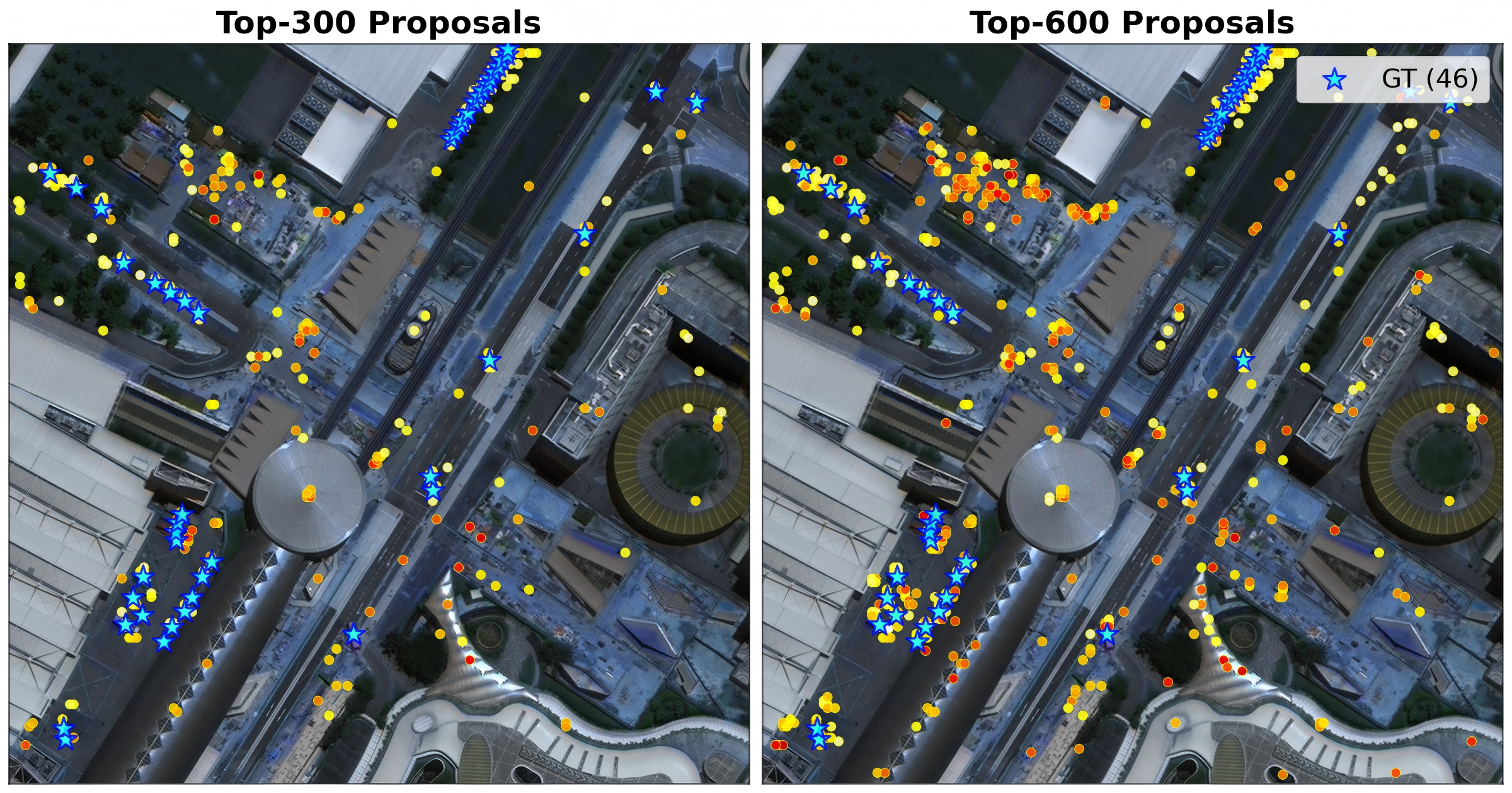}
    \label{fig:proposal_comparison_0186}
\end{subfigure}
\vspace{-12pt}
\caption{Qualitative examples of encoder proposal redundancy using different budgets.
Cyan stars denote GT centers, and proposal points are colored from yellow to red by confidence.
Even with Top-300, most GT objects are already covered by nearby proposals; increasing the budget to Top-600 mainly increases proposal density and redundancy rather than introducing new object coverage.
This redundancy helps explain why budget-matched Top-$K$ can closely match PAQI in overall AP.}
\label{fig:topk_redundancy_examples}
\end{figure}

\textbf{Redundancy.}
High overlap alone does not fully explain the similarity in overall AP.
Another contributing factor is the intrinsic redundancy of encoder proposals.
As shown in Fig.~\ref{fig:topk_redundancy_examples}, even under different budgets, GT objects are typically already surrounded by nearby proposals, while increasing $K$ mainly makes these proposals denser rather than introducing entirely new object coverage.
This suggests that encoder proposals often provide multiple candidate locations for the same object, especially around foreground regions.
Therefore, once the decoder receives a sufficiently large set of high-confidence proposals, increasing the budget further mainly adds redundant evidence rather than qualitatively new information.

Taken together, these observations help explain why a plain Top-$K$ initializer with sufficient capacity can closely reproduce PAQI's overall AP.
They support the budgeting interpretation without establishing that arbitrary query identities are interchangeable or that selection is unnecessary in all settings.

\section{Full Density-Stratified Results}
\label{app:density_stratified}

\begin{figure}[!t]
\centering
\captionsetup{width=0.98\linewidth, font=small}
\begin{subfigure}[t]{0.46\linewidth}
    \centering
    \includegraphics[width=\linewidth]{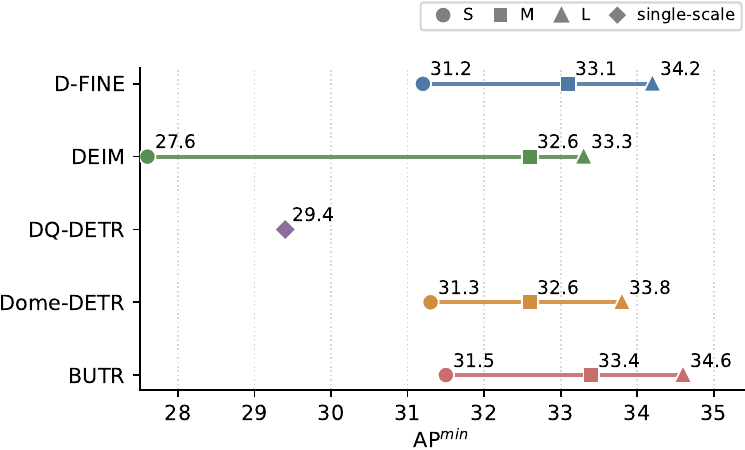}
    \vspace{-17pt}
    \caption{AI-TOD-v2: \texttt{minimal} subset}
    \label{fig:aitodv2_density_sparse}
\end{subfigure}
\hfill
\begin{subfigure}[t]{0.46\linewidth}
    \centering
    \includegraphics[width=\linewidth]{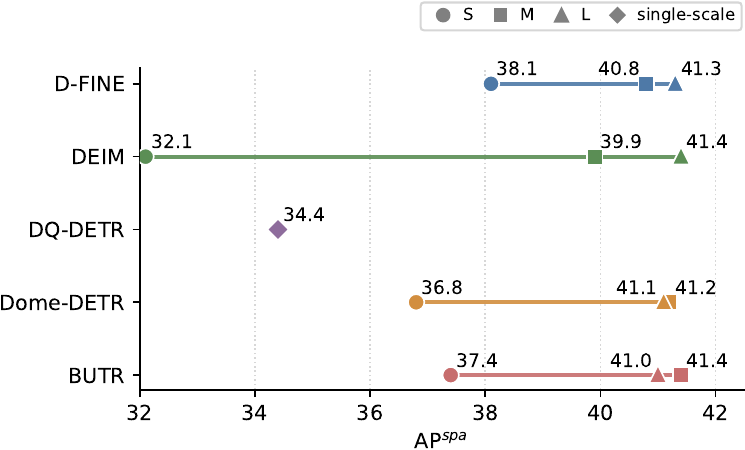}
    \vspace{-17pt}
    \caption{AI-TOD-v2: \texttt{sparse} subset}
    \label{fig:aitodv2_density_medium}
\end{subfigure}

\vspace{2pt}

\begin{subfigure}[t]{0.46\linewidth}
    \centering
    \includegraphics[width=\linewidth]{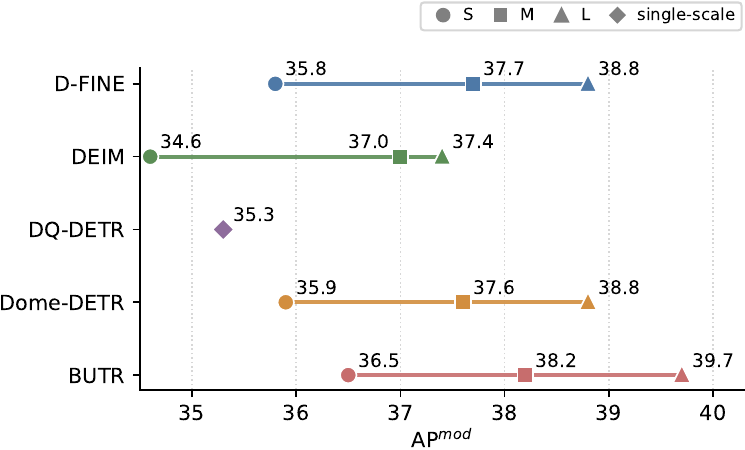}
    \vspace{-17pt}
    \caption{AI-TOD-v2: \texttt{moderate} subset}
    \label{fig:aitodv2_density_dense}
\end{subfigure}
\hfill
\begin{subfigure}[t]{0.46\linewidth}
    \centering
    \includegraphics[width=\linewidth]{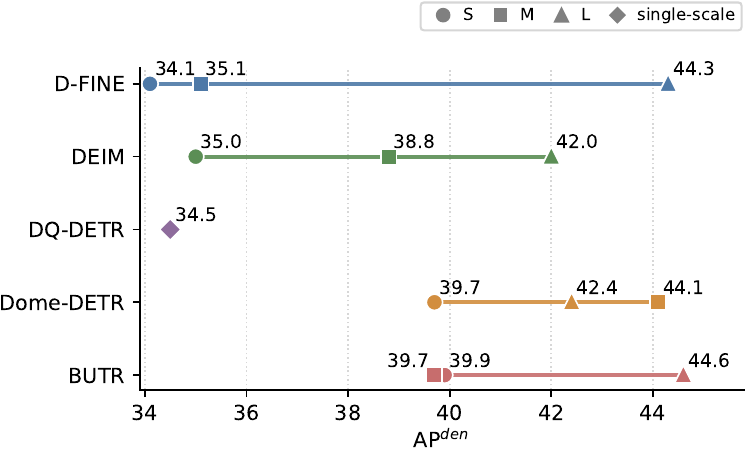}
    \vspace{-17pt}
    \caption{AI-TOD-v2: \texttt{dense} subset}
    \label{fig:aitodv2_density_extreme}
\end{subfigure}

\vspace{2pt}

\begin{subfigure}[t]{0.46\linewidth}
    \centering
    \includegraphics[width=\linewidth]{aitodv2_extreme_dotplot.pdf}
    \vspace{-17pt}
    \caption{AI-TOD-v2: \texttt{extreme} subset}
    \label{fig:aitodv2_extreme_dotplot2}
\end{subfigure}

\caption{Full density-stratified results on AI-TOD-v2, including the main-text \texttt{extreme} subset.}
\label{fig:aitodv2_full_density_stratified}
\end{figure}

\begin{figure}[!t]
\centering
\captionsetup{width=0.98\linewidth, font=small}

\begin{subfigure}[t]{0.46\linewidth}
    \centering
    \includegraphics[width=\linewidth]{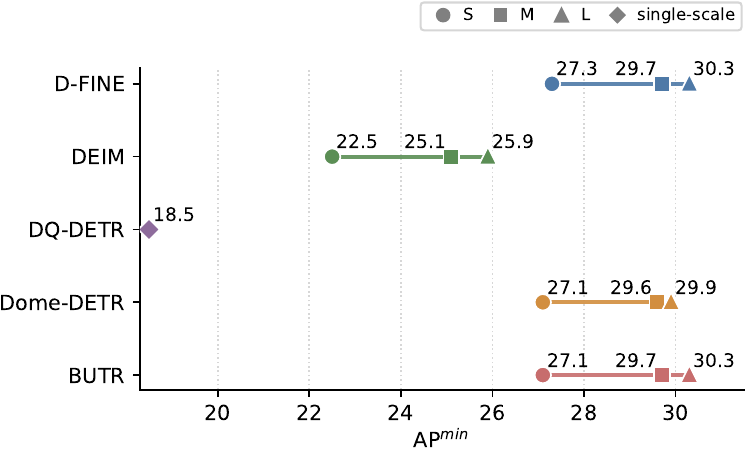}
    \vspace{-17pt}
    \caption{VisDrone: \texttt{minimal} subset}
    \label{fig:visdrone_density_sparse}
\end{subfigure}
\hfill
\begin{subfigure}[t]{0.46\linewidth}
    \centering
    \includegraphics[width=\linewidth]{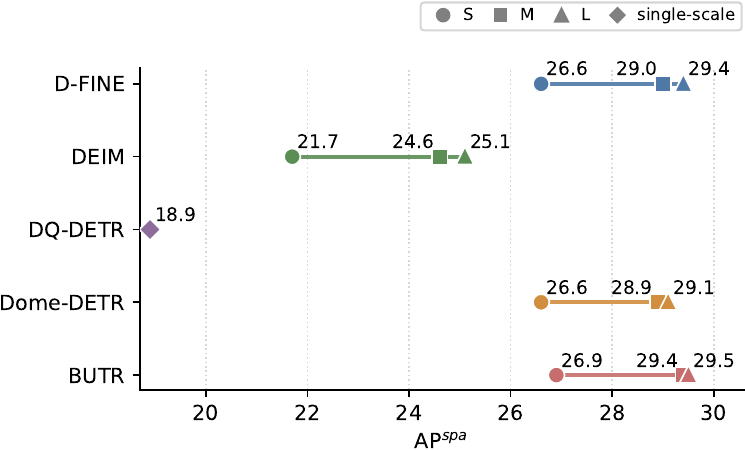}
    \vspace{-17pt}
    \caption{VisDrone: \texttt{sparse} subset}
    \label{fig:visdrone_density_medium}
\end{subfigure}

\vspace{2pt}

\begin{subfigure}[t]{0.46\linewidth}
    \centering
    \includegraphics[width=\linewidth]{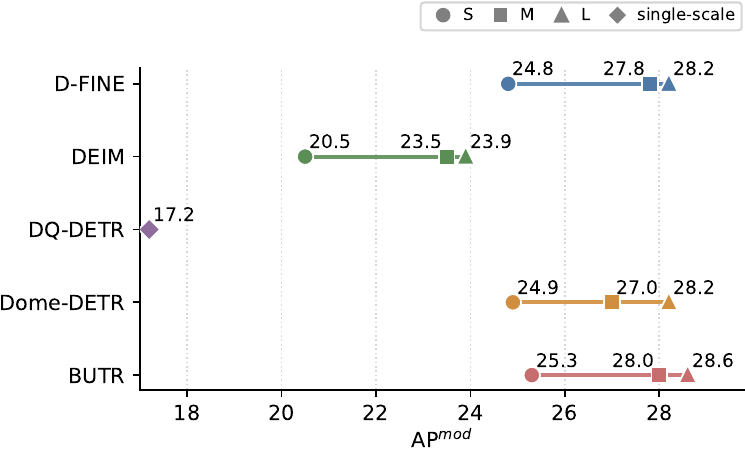}
    \vspace{-17pt}
    \caption{VisDrone: \texttt{moderate} subset}
    \label{fig:visdrone_density_dense}
\end{subfigure}
\hfill
\begin{subfigure}[t]{0.46\linewidth}
    \centering
    \includegraphics[width=\linewidth]{visdrone_dense_dotplot.pdf}
    \vspace{-17pt}
    \caption{VisDrone: \texttt{dense} subset}
    \label{fig:visdrone_dense_dotplot2}
\end{subfigure}

\caption{Full density-stratified results on VisDrone, including the main-text \texttt{dense} subset.}
\label{fig:visdrone_full_density_stratified}
\end{figure}

To complement Sec.~\ref{sec:density_stratified}, we provide the full density-stratified results in Fig.~\ref{fig:aitodv2_full_density_stratified} and Fig.~\ref{fig:visdrone_full_density_stratified}, including the tail subsets already reported in the main text.
BUTR's gains are observed in multiple lower-density subsets as well as the tail, although it does not outperform every baseline in every model-scale--subset comparison. These results describe performance across the density distribution rather than a decomposition of the overall AP gain.

\section{Stability of Tail Performance Across Training Runs}
\label{app:training_stability}

To assess the stability of performance on the small high-density tail, we conduct two complementary evaluations of Dome-DETR-L and BUTR-L on AI-TOD-v2 under matched training and evaluation settings.
We report AP$^{ext}$ on the same \texttt{extreme} subset across runs.

\textbf{Evaluation protocols.}
The main results use the predefined seed 0.
For fixed-seed reproducibility, we perform three independent training runs with seed 0, including the originally reported run.
For multi-seed robustness, we compare the two methods using matched seeds 0, 1, and 2; the seed-0 entry is the originally reported run and is shared with the fixed-seed evaluation.
For each run, we select the checkpoint with the highest overall AP and evaluate it separately on the \texttt{extreme} subset to obtain AP$^{ext}$.
Both evaluations report the mean and standard deviation over three runs.

\begin{table}[htbp]
\centering
\fontsize{8.2}{8.8}\selectfont
\captionsetup{width=0.98\linewidth, font=small}
\setlength{\tabcolsep}{6pt}
\renewcommand{\arraystretch}{1.1}
\caption{Fixed-seed reproducibility of AP$^{ext}$ on AI-TOD-v2: three independent seed-0 training runs. Run 1 is the originally reported result. Mean and standard deviation describe variation across training runs.}
\vspace{2pt}
\label{tab:fixed_seed_stability}
\begin{tabular}{lcccc}
\toprule
\textbf{Method} & \textbf{Run 1} & \textbf{Run 2} & \textbf{Run 3} & \textbf{AP}$^{ext}$ \\
\midrule
Dome-DETR-L & 29.9 & 29.1 & 28.8 & $29.27\pm0.57$ \\
\textbf{BUTR-L} & 29.8 & 29.7 & 29.4 & $\bm{29.63\pm0.21}$ \\
\bottomrule
\end{tabular}
\end{table}

\textbf{Fixed-seed reproducibility.}
As shown in Tab.~\ref{tab:fixed_seed_stability}, BUTR-L exhibits lower AP$^{ext}$ variation under the original seed-0 protocol: $29.63\pm0.21$, compared with $29.27\pm0.57$ for Dome-DETR-L.

\begin{table}[htbp]
\centering
\fontsize{8.2}{8.8}\selectfont
\captionsetup{width=0.98\linewidth, font=small}
\setlength{\tabcolsep}{6pt}
\renewcommand{\arraystretch}{1.1}
\caption{Multi-seed robustness of AP$^{ext}$ on AI-TOD-v2 using matched seeds 0, 1, and 2. Seed 0 is the originally reported result. Mean and sample standard deviation describe variation across seeds.}
\vspace{2pt}
\label{tab:multi_seed_stability}
\begin{tabular}{lcccc}
\toprule
\textbf{Method} & \textbf{Seed 0} & \textbf{Seed 1} & \textbf{Seed 2} & \textbf{AP}$^{ext}$ \\
\midrule
Dome-DETR-L & 29.9 & 27.2 & 27.5 & $28.20\pm1.48$ \\
\textbf{BUTR-L} & 29.8 & 29.3 & 29.3 & $\bm{29.47\pm0.29}$ \\
\bottomrule
\end{tabular}
\end{table}

\textbf{Multi-seed robustness.}
Tab.~\ref{tab:multi_seed_stability} shows that BUTR-L achieves higher mean AP$^{ext}$ with lower standard deviation ($29.47\pm0.29$ versus $28.20\pm1.48$).

Together, these results show lower observed variability in BUTR-L's tail performance under both fixed-seed repetitions and different random seeds in this configuration.
In light of the erroneous-filtering examples in App.~\ref{app:mask_filtering}, this stability advantage may partly stem from avoiding errors introduced by Dome-DETR's learned spatial mask: BUTR instead uses a non-learned budgeter to adjust the query count while retaining standard confidence-ranked initialization.

\section{Supplementary Experiment on Dense Agricultural Imagery}
\label{app:sorghum_protocol}
As a supplementary cross-domain evaluation, we use Sorghum-Grain-Count~\cite{bakshi2025sorghum}, which contains many small, visually similar grain instances outside the aerial domain.

\textbf{Data preparation.}
The original training and validation splits are combined into a training set, while the original test split is reserved for evaluation. We crop the $3456\times5184$ source images into overlapping $800\times800$ patches with a stride of 640 and retain patches containing at least one annotated grain.
Tab.~\ref{tab:sorghum_statistics} summarizes the resulting cropped splits. Annotation totals and objects-per-patch statistics refer to the retained patches, rather than counts of unique grains across the original images. This protocol evaluates detection on nonempty crops; it does not establish performance on empty-background images or full-image counting after merging overlapping predictions.

\begin{table}[htbp]
\centering
\fontsize{8.2}{8.8}\selectfont
\captionsetup{width=0.98\linewidth, font=small}
\setlength{\tabcolsep}{3pt}
\renewcommand{\arraystretch}{1.1}
\caption{Statistics of the cropped Sorghum-Grain-Count splits. Objects/patch reports the mean and standard deviation across retained patches. P90, P95, and P99 denote percentiles of the patch-level object-count distribution.}
\vspace{2pt}
\label{tab:sorghum_statistics}
\begin{tabular}{lrrrrrrrr}
\toprule
\textbf{Split} & \textbf{Patches} & \textbf{Annotations} & \textbf{Objects/patch} & \textbf{Median} & \textbf{Min--Max} & \textbf{P90} & \textbf{P95} & \textbf{P99} \\
\midrule
Trainval & 1,859 & 178,238 & $95.88\pm89.32$ & 70.0 & 1--429 & 229.0 & 272.0 & 350.4 \\
Test & 433 & 45,341 & $104.71\pm94.31$ & 75.0 & 1--523 & 236.8 & 277.4 & 361.3 \\
\bottomrule
\end{tabular}
\end{table}

\textbf{Comparison protocol and results.}
Within this comparison, all models use matched training and evaluation settings with $800\times800$ inputs. We evaluate D-FINE-L and DEIM-L with fixed budgets of 400 and 1100 queries, close to the observed average query counts of BUTR-L (378) and Dome-DETR-L (1103), respectively.
As shown in Tab.~\ref{tab:sorghum_results}, raising the fixed budget from 400 to 1100 improves AP from 57.0 to 57.7 for D-FINE-L and from 57.2 to 58.0 for DEIM-L. BUTR-L achieves 58.0 AP with an average of only 378 queries, exceeding D-FINE-L and matching DEIM-L at the larger budget while requiring less training time. It also improves over Dome-DETR-L by 0.4 AP while using approximately one-third as many queries, with $2.2\times$ faster training.
These results provide supplementary evidence for confidence-based budgeting in a dense agricultural setting.

\begin{table}[htbp]
\centering
\fontsize{8.2}{8.8}\selectfont
\captionsetup{width=0.98\linewidth, font=small}
\setlength{\tabcolsep}{3.5pt}
\renewcommand{\arraystretch}{1.1}
\caption{Comparison on Sorghum-Grain-Count using the cropped trainval/test protocol. Query-count mean and standard deviation describe variation across evaluation patches, not repeated training runs. Tr. (h): training time. T4 deployment latency was not measured for this supplementary experiment.}
\vspace{2pt}
\label{tab:sorghum_results}
\begin{tabular}{l|c|lc|cccccc}
\toprule
\textbf{Method} & \textbf{Tr. (h)} & \textbf{Init.} & $|Q|$ ($\mu\pm\sigma$) & \textbf{AP} & \textbf{AP}$_{50}$ & \textbf{AP}$_{75}$ & \textbf{AP}$_t$ & \textbf{AP}$_s$ & \textbf{AP}$_m$ \\
\midrule
D-FINE-L & 1.5 & Fixed & 400 & 57.0 & 92.6 & 64.9 & 10.6 & 50.1 & 63.6 \\
D-FINE-L & 1.7 & Fixed & 1100 & 57.7 & 93.9 & 65.6 & 13.4 & 50.9 & 64.2 \\
DEIM-L & 1.6 & Fixed & 400 & 57.2 & 92.9 & 65.1 & 10.9 & 50.3 & 63.8 \\
DEIM-L & 1.8 & Fixed & 1100 & \textbf{58.0} & 94.0 & 66.3 & 14.3 & 51.3 & 64.3 \\
Dome-DETR-L & 3.3 & Dynamic & $1103\pm195$ & 57.6 & 94.0 & 65.4 & 13.1 & 50.8 & 64.1 \\
\midrule
\textbf{BUTR-L} & 1.5 & Dynamic & $378\pm119$ & \textbf{58.0} & 93.8 & 66.1 & 14.5 & 51.5 & 64.1 \\
\bottomrule
\end{tabular}
\end{table}

\section{T4 Deployment Latency Protocol}
\label{app:t4_deployment}

\paragraph{Measurement protocol.}
We use a single NVIDIA Tesla T4 (16\,GB) on Kaggle, Python 3.10.14, PyTorch 2.4.0+cu118, and TensorRT 10.4.0, with batch size 1. Inputs are $800\times800$ for tiny-object benchmarks and $640\times640$ for COCO. Paired comparisons run sequentially on the same physical GPU. Each result averages three rounds, each comprising 1,000 warmup calls and the same ordered 1,000-image subset of the corresponding test split (val2017 for COCO); all samples and rounds are retained.
We use the D-FINE/DEIM synchronized timing boundary with synchronous \texttt{execute\_v2}, without CUDA Graphs. The timer includes all model stages, detection post-processing, and dynamic-budget computation, readback, synchronization, and shape handling. Image loading, preprocessing, and input transfer are excluded.

\paragraph{Precision and layout.}
Tiny-object engines enable FP16 with sensitive operations constrained to FP32; DQ-DETR additionally protects bounding-box and sampling-coordinate operations. TF32 is disabled. COCO uses the official default \texttt{trtexec --fp16} build settings. BUTR uses separate encoder/decoder engines and splits cross-attention value features by spatial scale before transposing each scale. This layout rewrite preserves the budgeting rule and decoder mathematics. D-FINE/DEIM retain their original single-engine graphs; Dome-DETR and DQ-DETR use front/back engines, with all stages included in timing.

\paragraph{Layout fairness controls.}
Applying the same rewrite to complete AI-TOD-v2 D-FINE-S/DEIM-S graphs increases latency from 12.14/12.23 ms for independently rebuilt original graphs to 15.17/15.32 ms. We therefore retain the faster original baseline layouts. These matched controls show that the rewrite's benefit depends on the surrounding graph and engine boundaries, rather than establishing a universal speedup.

\paragraph{COCO calibration.}
For official D-FINE-S/M/L/X checkpoints with 300 queries, we measure 3.624/5.635/7.909/13.485 ms, versus published values of 3.49/5.62/8.07/12.89 ms; all deviations are within 5\%. The physical T4 is selected using the predeclared mean absolute relative deviation across all four calibration means (2.68\%), without using BUTR performance. Paired COCO plug-in measurements use this same card.

\paragraph{Accuracy scope.}
AP is reported from the detection experiments, rather than a full-test FP16 deployment evaluation. BUTR's paired layout checks preserve per-image budgets and change subset AP by less than 0.04 points across both tiny-object datasets. AI-TOD-v2 DEIM-S timing remains provisional because of an unresolved deployment-accuracy mismatch. VisDrone DEIM-S/M uses same-scale D-FINE weights with the DEIM inference architecture. Full deployment scripts and additional checks are provided in the repository.

\section{Broader Impact}
The broader impact of this work is primarily methodological.
By reframing dynamic query as a budgeting problem, we aim to clarify the mechanism behind adaptive query allocation and encourage simpler, more principled designs in tiny-object detection.
Our results also suggest that density-stratified evaluation can serve as a useful complementary protocol when overall AP obscures behavior on rare but important high-density cases.
Beyond tiny-object detection, the COCO plug-in results suggest that simple query budgeting can also be useful in general-purpose detection.

At the application level, improving efficiency and reliability in dense and tiny-object scenarios may benefit tasks such as remote sensing, traffic analysis, and environmental monitoring.
At the same time, stronger detection in aerial or crowded scenes could also be misused for intrusive surveillance or monitoring, so downstream deployment should still consider appropriate privacy, legal, and ethical safeguards.

\newpage
\clearpage
\section*{NeurIPS Paper Checklist}

\begin{enumerate}

\item {\bf Claims}
    \item[] Question: Do the main claims made in the abstract and introduction accurately reflect the paper's contributions and scope?
    \item[] Answer: \answerYes{} 
    \item[] Justification: The abstract and introduction state the paper's main contributions: a mechanism-oriented diagnosis of dynamic query, the budgeting view, the proposed BUTR framework, and evaluations on AI-TOD-v2, VisDrone, and COCO. The claims are supported by the diagnostic analyses, ablations, and experimental results reported in the paper.

\item {\bf Limitations}
    \item[] Question: Does the paper discuss the limitations of the work performed by the authors?
    \item[] Answer: \answerYes{} 
    \item[] Justification: The paper discusses the scope and limitations of the approach, including its focus on two-stage DETR-style detectors, the use of encoder confidences as a density proxy, and the empirical nature of the evaluation across the considered datasets and baselines.

\item {\bf Theory assumptions and proofs}
    \item[] Question: For each theoretical result, does the paper provide the full set of assumptions and a complete (and correct) proof?
    \item[] Answer: \answerNA{} 
    \item[] Justification: The paper does not present theoretical results, theorems, or formal proofs. The proposed budgeting principle is empirically motivated and evaluated through experiments.

    \item {\bf Experimental result reproducibility}
    \item[] Question: Does the paper fully disclose all the information needed to reproduce the main experimental results of the paper to the extent that it affects the main claims and/or conclusions of the paper (regardless of whether the code and data are provided or not)?
    \item[] Answer: \answerYes{} 
    \item[] Justification: The paper specifies the datasets, splits, evaluation protocol, model variants, training schedules, query budgets, latency measurement protocol, hardware, and implementation details needed to reproduce the main results. The BUTR procedure is also described algorithmically in the appendix.

\item {\bf Open access to data and code}
    \item[] Question: Does the paper provide open access to the data and code, with sufficient instructions to faithfully reproduce the main experimental results, as described in supplemental material?
    \item[] Answer: \answerYes{} 
    \item[] Justification: The experiments are conducted on publicly available datasets and based on publicly available detector implementations/checkpoints. We provide code and instructions for reproducing the main experiments in the supplementary material.

\item {\bf Experimental setting/details}
    \item[] Question: Does the paper specify all the training and test details (e.g., data splits, hyperparameters, how they were chosen, type of optimizer) necessary to understand the results?
    \item[] Answer: \answerYes{}
    \item[] Justification: The paper reports the datasets and splits, input resolution, training schedules, query budgets, evaluation settings, evaluation protocol, latency measurement protocol, and hardware configuration.

\item {\bf Experiment statistical significance}
    \item[] Question: Does the paper report error bars suitably and correctly defined or other appropriate information about the statistical significance of the experiments?
    \item[] Answer: \answerNo{}
    \item[] Justification: The main detection results are reported as standard single-run COCO-style AP evaluations, following common practice. For diagnostic analyses where applicable, the paper reports mean and standard deviation over images, such as query counts, overlap ratios, and query-usage behavior.

\item {\bf Experiments compute resources}
    \item[] Question: For each experiment, does the paper provide sufficient information on the computer resources (type of compute workers, memory, time of execution) needed to reproduce the experiments?
    \item[] Answer: \answerYes{}
    \item[] Justification: The paper reports the compute platform, including CPUs, GPUs, input resolution, batch size, training-time measurement protocol, and per-image inference latency. Training wall-clock time and inference latency are reported for the main comparisons.
    
\item {\bf Code of ethics}
    \item[] Question: Does the research conducted in the paper conform, in every respect, with the NeurIPS Code of Ethics \url{https://neurips.cc/public/EthicsGuidelines}?
    \item[] Answer: \answerYes{}
    \item[] Justification: The research uses public datasets and standard object detection benchmarks, and does not involve deceptive data collection, human-subject experiments, or privacy-invasive data collection by the authors. We have reviewed the NeurIPS Code of Ethics and believe the work conforms to it.

\item {\bf Broader impacts}
    \item[] Question: Does the paper discuss both potential positive societal impacts and negative societal impacts of the work performed?
    \item[] Answer: \answerYes{}
    \item[] Justification: The paper discusses potential positive impacts, such as improving the efficiency and reliability of tiny-object detection, as well as potential negative impacts, including misuse of improved aerial or dense-scene detection systems for surveillance or monitoring. The method is a general detection technique and is not deployed in any real-world system in this work.
    
\item {\bf Safeguards}
    \item[] Question: Does the paper describe safeguards that have been put in place for responsible release of data or models that have a high risk for misuse (e.g., pre-trained language models, image generators, or scraped datasets)?
    \item[] Answer: \answerNA{}
    \item[] Justification: The paper does not release high-risk generative models, scraped datasets, or models specifically designed for misuse. The work uses existing public detection datasets and focuses on algorithmic analysis and efficiency.

\item {\bf Licenses for existing assets}
    \item[] Question: Are the creators or original owners of assets (e.g., code, data, models), used in the paper, properly credited and are the license and terms of use explicitly mentioned and properly respected?
    \item[] Answer: \answerYes{}
    \item[] Justification: The paper cites the datasets, baseline methods, and detector implementations/checkpoints used in the experiments. Existing assets are used according to their respective licenses and terms of use.

\item {\bf New assets}
    \item[] Question: Are new assets introduced in the paper well documented and is the documentation provided alongside the assets?
    \item[] Answer: \answerYes{}
    \item[] Justification: The paper introduces code for the proposed BUTR budgeter and evaluation scripts. The released assets are documented with instructions for reproducing the main experiments and are anonymized for submission when applicable.

\item {\bf Crowdsourcing and research with human subjects}
    \item[] Question: For crowdsourcing experiments and research with human subjects, does the paper include the full text of instructions given to participants and screenshots, if applicable, as well as details about compensation (if any)? 
    \item[] Answer: \answerNA{}
    \item[] Justification: The paper does not involve crowdsourcing, user studies, or research with human subjects.

\item {\bf Institutional review board (IRB) approvals or equivalent for research with human subjects}
    \item[] Question: Does the paper describe potential risks incurred by study participants, whether such risks were disclosed to the subjects, and whether Institutional Review Board (IRB) approvals (or an equivalent approval/review based on the requirements of your country or institution) were obtained?
    \item[] Answer: \answerNA{}
    \item[] Justification: The paper does not involve crowdsourcing or research with human subjects, so IRB approval or equivalent review is not applicable.

\item {\bf Declaration of LLM usage}
    \item[] Question: Does the paper describe the usage of LLMs if it is an important, original, or non-standard component of the core methods in this research? Note that if the LLM is used only for writing, editing, or formatting purposes and does \emph{not} impact the core methodology, scientific rigor, or originality of the research, declaration is not required.
    \item[] Answer: \answerNA{}
    \item[] Justification: The core method development, experiments, and scientific claims do not involve LLMs as an important, original, or non-standard component. Any LLM use, if any, was limited to writing, editing, or formatting assistance and did not affect the methodology, scientific rigor, or originality of the research.

\end{enumerate}

\end{document}